\documentclass[final,5p,times,hidelinks,twocolumn]{elsarticle}
\usepackage[hidelinks]{hyperref}

\journal{Medical Image Analysis}

\makeatletter
\def\ps@pprintTitle{%
 \let\@oddhead\@empty
 \let\@evenhead\@empty
 \def\@oddfoot{}%
 \let\@evenfoot\@oddfoot}
\makeatother

\biboptions{authoryear}

\usepackage{amsmath}
\usepackage{amssymb}
\usepackage{amsmath,mathtools}
\usepackage{array}
\usepackage{caption}
\usepackage{color}
\usepackage{float}
\usepackage{graphicx}
\usepackage{hhline}
\usepackage[utf8x]{inputenc}
\usepackage{longtable}
\usepackage{multirow}
\usepackage{rotating}
\usepackage{setspace}
\usepackage{subcaption}
\usepackage{tabularx}
\usepackage{url}
\usepackage{tablefootnote}

\usepackage[percent]{overpic}
\usepackage{pict2e}

\newcolumntype{Y}{>{\RaggedRight\arraybackslash}X}

\usepackage{array}
\newcolumntype{M}[1]{>{\centering\arraybackslash\hspace{0pt}}m{#1}}
\newcolumntype{P}[1]{>{\centering\arraybackslash}p{#1}}

\begin{document}

\begin{frontmatter}

\title{Quantitative analysis of patch-based fully convolutional neural networks for tissue segmentation on brain magnetic resonance imaging}

\author[]{Jose Bernal\corref{mycorrespondingauthor}}
\cortext[mycorrespondingauthor]{Corresponding author}
\ead{jose.bernal@udg.edu}

\author[]{Kaisar Kushibar}
\ead{kaisar.kushibar@udg.edu}

\author[]{Mariano Cabezas}
\ead{mariano.cabezas@udg.edu}

\author[]{Sergi Valverde}
\ead{sergi.valverde@udg.edu}

\author[]{Arnau Oliver}
\ead{arnau.oliver@udg.edu}

\author[]{Xavier Llad\'o}
\ead{xavier.llado@udg.edu}

\address[mymainaddress]{Computer Vision and Robotics Institute\\
    Dept. of Computer Architecture and Technology\\
    University of Girona\\
    Ed. P-IV, Av. Lluis Santal\'o s/n, 17003 Girona (Spain)}

\begin{abstract}
Accurate brain tissue segmentation in Magnetic Resonance Imaging (MRI) has attracted the attention of medical doctors and researchers since variations in tissue volume help in diagnosing and monitoring neurological diseases. Several proposals have been designed throughout the years comprising conventional machine learning strategies as well as convolutional neural networks (CNN) approaches. In particular, in this paper, we analyse a sub-group of deep learning methods producing dense predictions. This branch, referred in the literature as Fully CNN (FCNN), is of interest as these architectures can process an input volume in less time than CNNs and local spatial dependencies may be encoded since several voxels are classified at once. Our study focuses on understanding architectural strengths and weaknesses of literature-like approaches. Hence, we implement eight FCNN architectures inspired by robust state-of-the-art methods on brain segmentation related tasks. We evaluate them using the IBSR18, MICCAI2012 and iSeg2017 datasets as they contain infant and adult data and exhibit varied voxel spacing, image quality, number of scans and available imaging modalities. The discussion is driven in three directions: comparison between 2D and 3D approaches, the importance of multiple modalities and overlapping as a sampling strategy for training and testing models. To encourage other researchers to explore the evaluation framework, a public version is accessible to download from our research website.

\end{abstract}


\end{frontmatter}


\section{Introduction\label{sec:intro}}
Automatic brain Magnetic Resonance Imaging (MRI) tissue segmentation continues being an active research topic in medical image analysis since it provides doctors with meaningful and reliable quantitative information, such as tissue volume measurements. This information is widely used to diagnose brain diseases and to evaluate progression through regular MRI analysis over time~\citep{rovira2015evidence, steenwijk2016cortical, Filippi2016}. Hence, MRI and its study contribute to a better comprehension of the nature of brain problems and the effectiveness of new treatments.

Several tissue segmentation algorithms have been proposed throughout the years. Many supervised machine learning methods existed before the Convolutional Neural Network (CNN) era. A clear example of that is the pipelines that participated in the MRBrainS13 challenge~\citep{mendrik2015mrbrains}. Commonly, intensity-based methods assumed each tissue could be represented by its intensity values~\citep{cardoso2013adapt} (e.g. using GMM models). Since noise and intensity inhomogeneities degraded them, they were later equipped with spatial information~\citep{clarke1995mri, kapur1996segmentation, liew2006current,valverde2015comparison}. Four main strategies were distinguished in the literature: (i) impose local contextual constraints using Markov Random Fields (MRF)~\citep{FAST2001}, (ii) include penalty terms accounting for neighbourhood similarity in clustering objective functions~\citep{FANTASM2001}, (iii) use Gibbs prior to model spatial characteristics of the brain~\citep{PVC2001} and (iv) introduce spatial information using probabilistic atlases~\citep{SPM5, SMP8}. It is important to remark that some of these methods, like FAST~\citep{FAST2001} and SPM~\citep{SPM5, SMP8}, are still being used in medical centres due to their robustness and adaptability~\citep{valverde2017automated}.

Nowadays, CNNs have become appealing to address this task in coming years since (i) they have achieved record-shattering performances in various fields in computer vision and (ii) they discover classification-suitable representations directly from the input data -- unlike conventional machine-learning strategies. However, unlike traditional approaches, these methods still present two main issues when placed in real life scenarios: (i) lack of sufficiently labelled data and (ii) domain adaptation issues -- also related to generalisation problems. Seminal work on CNN for brain tissue segmentation date back to $2015$ when~\cite{Zhang2015} proposed a CNN to address infant brain tissue segmentation on MRI where tissue distributions overlap and, hence, the GMM assumption does not hold. The authors showed that their CNN was suitable for the problem and could outperform techniques, such as random forest, support vector machines, coupled level sets, and majority voting. From thereon, many more sophisticated proposals have been devised~\citep{litjens2017survey, bernal2017review}.

Former CNN strategies for tissue segmentation were trained to provide a single label given an input patch~\citep{Zhang2015, moeskops2016automatic, chen2017}. Naturally, both training and testing can be time-consuming and computationally demanding. Also, the relationship between neighbouring segmented voxels is not encoded -- in principle -- and, consequently, additional components on the architecture (such as in~\citep{LSTM2015}) or post-processing may be needed to smooth results. These drawbacks can be diminished by adapting the network to perform dense prediction. The prevailing approach consists in replacing fully connected layers by $1\times 1$ convolutional layers  -- $1\times 1\times 1$ if processing 3D data. This particular group is known as Fully CNN (FCNN)~\citep{long2015fully}. 

Regarding input dimensionality, three main streams are identified: 2D, 2.5D and 3D. At the beginning of the CNN era, most of the state-of-the-art CNN techniques were 2D, in part, due to (i) their initial usage on natural images, and (ii) computation limitations of processing 3D volumes directly. Surely, three independent 2D models can be arranged to handle patches from axial, sagittal and coronal at the same time, hence improving acquired contextual information. These architectures are referred in the literature as 2.5D~\citep{lyksborg2015ensemble, birenbaum2016longitudinal, kushibar2017automated}. With advances in technology, more 3D approaches have been developed and attracted more researchers as they tend to outperform 2D architectures~\citep{bernal2017review}. Intuitively, the improvement of 3D over 2D and 2.5D lies on the fact that more information from the three orthogonal planes is integrated into the network -- i.e., more contextual knowledge is acquired. However, this does not hint they always perform better~\citep{ghafoorian2016location}. It is important to highlight that, to the best of our knowledge, no 2.5D FCNN network has been created yet.

In this paper, we analyse quantitatively $4\times 2$ FCNN architectures for tissue segmentation on brain MRI. These networks, comprising 2D and 3D implementations, are inspired in four recent works~\citep{cciccek20163d, Dolz2017, guerrero2017white, kamnitsas2017efficient}. The models are tested on three well-known datasets of infant and adult brain scans, with different spatial resolution, voxel spacing, and image modalities. 
In this paper, we aim to (i) compare different FCNN strategies for tissue segmentation; (ii) quantitatively analyse the effect of network's dimensionality (2D or 3D) for tissue segmentation and the impact of fusing information from single or multiple modalities; and (iii) investigate the effects of extracting patches with a certain degree of overlap as a sampling strategy in both training and testing. To the best of our knowledge, this is the first work providing a comprehensive evaluation of FCNNs for the task mentioned above.

The rest of the paper is organised as follows. In Section~\ref{sec:method}, we present our evaluation framework: selected datasets, assessed networks, aspects to analyse, pipeline description and implementation details. Results are reported in Section~\ref{sec:results} and analysed in Section~\ref{sec:discussion}. Final remarks are discussed in Section~\ref{sec:conclusions}.

\section{Materials and methods \label{sec:method}}
\subsection{Considered datasets}
Public datasets are commonly used for assessing brain MRI tissue segmentation algorithms as they provide ground truth labels. In this work, we consider one publicly available repository and two challenges: Internet Brain Segmentation Repository 18 (IBSR18)\footnote{http://www.nitrc.org/projects/ibsr}, MICCAI Multi-Atlas Labeling challenge 2012 (MICCAI 2012)\footnote{https://masi.vuse.vanderbilt.edu/workshop2012} and 6-month infant brain MRI segmentation (iSeg2017)\footnote{http://iseg2017.web.unc.edu/}, respectively. The datasets were chosen since (i) they have been widely used in the literature to compare different methods and also (ii) they contain infants and adults data, with different voxel spacing and a varied number of scans. To use annotations of MICCAI 2012, we mapped all the labels to form the three tissue classes. We believe that these two factors allow us to see how robust, generalised and useful in different scenarios the algorithms can be. Specific details of these datasets are presented in Table~\ref{tab:dataset-information}.

\begin{table}
    \centering
    \caption{Relevant information from the considered datasets. In the table, the elements to be considered are presented in the first column and the corresponding information from IBSR18, MICCAI 2012 and iSeg2017 are detailed in the following ones. In the row related to the number of scans (with GT), the number of training and test volumes is separated by a $+$ sign. For both IBSR18 and iSeg2017, the evaluation is carried out using leave-one-out cross-validation.\label{tab:dataset-information}}
    {\footnotesize
    \begin{tabular}{|l|c|c|c|}
        \hline
        \textbf{Item}  & \textbf{IBSR18} & \textbf{MICCAI 2012} & \textbf{iSeg2017} \\ \hline
        Target & Adult & Adult & Infants \\ \hline
        Number of scans & $18$ & $15+20$ & $10$ \\ \hline
        Bias-field corrected & Yes & Yes & Yes \\ \hline
        Intensity corrected & No & Yes & No \\ \hline
        Skull stripped & No & No & Yes \\ \hline
        Voxel spacing & Anisotropic & Isotropic & Isotropic \\ \hline
        Modalities & T1-w & T1-w & T1-w, T2-w \\ \hline
    \end{tabular}}
\end{table}

\subsection{FCNNs for brain MRI segmentation tasks}
The proposed works using FCNN for brain MRI segmentation tasks are listed in Table~\ref{tab:articles}. Proposals comprise single or multiple flows of information -- referred in the literature as single-path and multi-path architectures, respectively. While single-path networks process input data faster than multi-path, knowledge fusion occurring in the latter strategy leads to better segmentation results: various feature maps from different interconnected modules and shallow layers are used to produce a final verdict~\citep{moeskops2016automatic}. Under this scheme, a network is provided with contrast, fine-grained and implicit contextual information. Furthermore, proposals apply successive convolutions only or convolutions and de-convolutions in the so-called u-shaped models. The latter approach commonly considers connections from high-resolution layers to up-sampled ones to retain location and contextual information~\citep{cciccek20163d,ronneberger2015u, milletari2016v}.

\begin{table}
    \centering
    \caption{Significant information of state-of-the-art FCNN approaches for brain segmentation tasks. The reference articles are listed in the first column. The following columns outline information regarding dimensionality of the input, high-level architectural details and segmentation problem addressed by the authors. U-shaped architectures are denoted by ``[U]". \label{tab:articles}}
    {\footnotesize
    \begin{tabular}{|l|l|l|}
    \hline
    \textbf{Article} & \textbf{Architecture} & \textbf{Target} \\ \hline
    \cite{brosch2016deep} & 3D multi-path [U] & Lesion \\ \hline
    \cite{kleesiek2016deep} & 3D single-path & Skull stripping \\ \hline
    \cite{nie2016fully} & 2D single-path [U] & Tissue \\ \hline
    \cite{shakeri2016sub} & 2D single-path & Sub-cortical structure\\ \hline
    \cite{kamnitsas2017efficient} & 3D multi-path & Lesion/tumour \\ \hline
    \cite{Dolz2017} & 3D multi-path & Sub-cortical structure  \\ \hline
    \cite{guerrero2017white} & 2D multi-path [U] & Lesion \\ \hline
    \cite{moeskops2017adversarial} & 2D single-path & Structure \\ \hline
    \end{tabular}}
\end{table}

From the papers indexed in Table~\ref{tab:articles}, we built four multi-path architectures inspired by the works of \cite{kamnitsas2017efficient}, \cite{Dolz2017}, \cite{cciccek20163d} and \cite{guerrero2017white} (i.e. two convolution-only and two u-shaped architectures). The networks were implemented in 2D and 3D to investigate the effect of the network's dimensionality in tissue segmentation. All these architectures were implemented from scratch following the architectural details given in the original work and are publicly available at our research website\footnote{https://github.com/NIC-VICOROB/tissue\_segmentation\_comparison}. Although slight architectural differences may be observed, the core idea of the proposals is retained. More details of the networks are given in the following sections.

\begin{figure*}
  \centering
  \bgroup
  \begin{tabular}{M{0.25\textwidth}M{0.7\textwidth}}
    \begin{subfigure}[b]{0.16\textwidth}
        \centering
        \includegraphics[width=\textwidth]{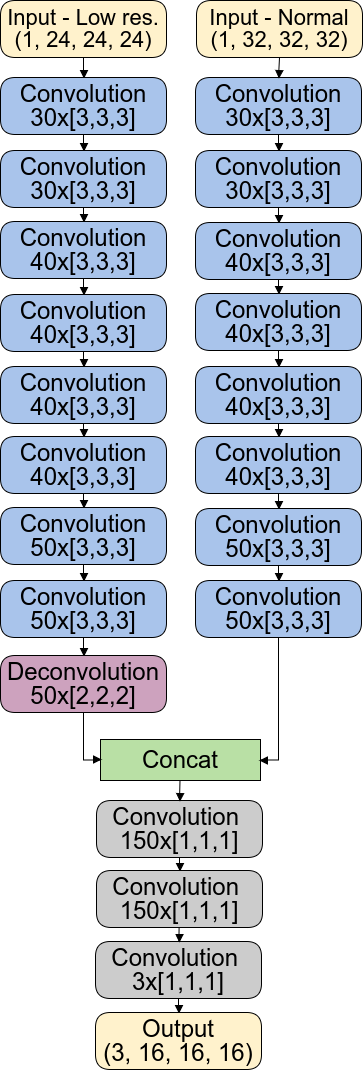}
        \caption{\label{fig:considered-networks-kk}}
    \end{subfigure}
   & 
    \subcaptionbox{\label{fig:considered-networks-dm}}{\includegraphics[width=0.62\textwidth]{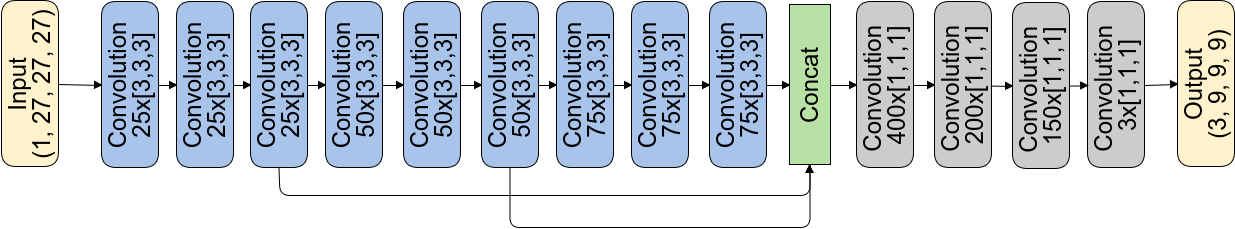}} \newline
    \subcaptionbox{\label{fig:considered-networks-u-shaped}}{\includegraphics[width=0.62\textwidth]{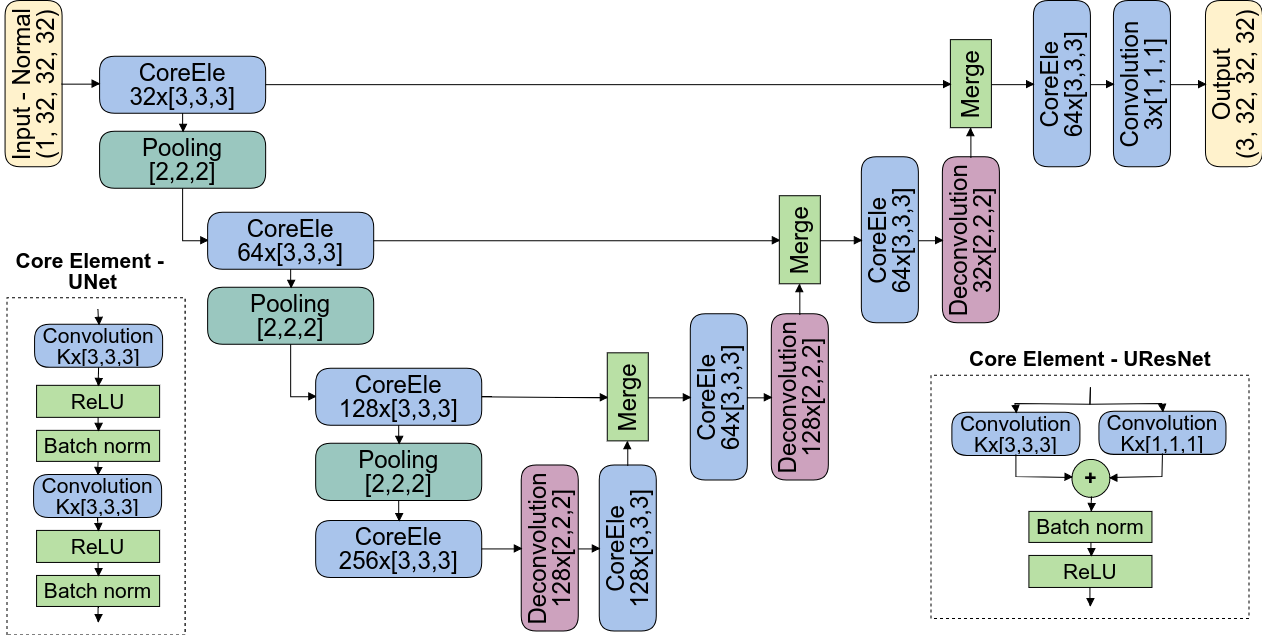}}
  \end{tabular}
  \egroup
  \caption{High level diagram of considered networks. Our implementations are inspired by the works of (a) \cite{kamnitsas2017efficient}, (b) \cite{Dolz2017}, and (c) \cite{cciccek20163d} and \cite{guerrero2017white}. Only 3D versions are shown. Notation is as follows: four-element tuples indicate number of channels and patch size in $x$, $y$ and $z$, in that order; triples in brackets indicate kernel size. In (c), merging is either concatenation or addition; CoreEle stands for core elements of the models (both of them are detailed on the bottom left and right corner of the (c)); the letter $K$ on the core elements is the number of kernels at a given stage.\label{fig:considered-networks}}
\end{figure*}

\subsubsection{Networks incorporating multi-resolution information}
\cite{kamnitsas2017efficient}, proposed a two-path 3D FCNN applied to brain lesion segmentation. This approach achieved top performance on two public benchmarks, BRATS 2015 and ISLES 2015. By processing information of the targeted area from two different scales simultaneously, the network incorporated local and larger contextual information, providing a more accurate response~\citep{moeskops2016automatic}. A high-level scheme of the architecture is depicted in Fig.~\ref{fig:considered-networks-kk}. Initially, two independent feature extractor modules extracted maps from patches from normal and downscaled versions of an input volume. Each module was formed by eight $3\times 3\times 3$ convolutional layers using between $30$ and $50$ kernels. Afterwards, features maps were fused and mined by two intermediate $1\times 1\times 1$ convolutional layers with $150$ kernels. Finally, a classification layer (another $1\times 1\times 1$ convolutional layer) produced segmentation prediction using a softmax activation.

\cite{Dolz2017} presented a multi-resolution 3D FCNN architecture applied to sub-cortical structure tissue segmentation. A general illustration of the architecture is shown in Fig.~\ref{fig:considered-networks-dm}. The network consisted of 13 convolutional layers: nine $3\times 3\times 3$, and four $1\times 1\times 1$. Each one of these layers was immediately followed by a Parametric Rectified Linear Unit (PReLU) layer, except for the output layer which activation was softmax. Multi-resolution information was integrated into this architecture by concatenating feature maps from shallower layers to the ones resulting from the last $3\times 3\times 3$ convolutional layer. As explained by~\cite{hariharan2015hypercolumns}, this kind of connections allows networks to learn semantic -- coming from deeper layers -- as well as fine-grained localisation information -- coming from shallow layers.

\subsubsection{U-shaped networks}
In the u-shaped network construction scheme, feature maps from higher resolution layers are commonly merged to the ones on deconvolved maps to keep localisation information. Merging has been addressed in the literature through concatenation~\citep{cciccek20163d,brosch2016deep} and addition~\citep{guerrero2017white}. In this paper, we consider networks using both approaches. A general scheme of our implementations inspired in both works is displayed in Fig.~\ref{fig:considered-networks-u-shaped}.

\cite{cciccek20163d} proposed a 3D u-shaped FCNN, known as 3D u-net. The network is formed by four convolution-pooling layers and four deconvolution-convolution layers. The number of kernels ranged from $32$ in its bottommost layers to $256$ in its topmost ones. In this design, maps from higher resolutions were concatenated to upsampled maps. Each convolution was immediately followed by a Rectified Linear Unit (ReLU) activation function.

\cite{guerrero2017white} designed a 2D u-shaped residual architecture applied on lesion segmentation, referred as u-ResNet. The building block of this network was the residual module which (i) added feature maps produced by $3\times 3$- and $1\times 1$-kernel convolution layers, (ii) normalised resulting features using batchnorm, and, finally, (iii) used a ReLU activation. The network consisted of three residual modules with $32$, $64$ and $128$ kernels, each one followed by a $2\times 2$ max pooling operation. Then, a single residual module with $256$ kernels was applied. Afterwards, successive deconvolution-and-residual-module pairs were employed to enlarge the networks' output size. The number of filters went from $256$ to $32$ in the layer before the prediction one. Maps from higher resolutions were merged with deconvolved maps through addition.

\subsection{Evaluation measurement}
To evaluate proposals, we used the Dice similarity coefficient (DSC)~\citep{dice1945measures, overlap2006}. The DSC is used to determine the extent of overlap between a given segmentation and the ground truth. Given an input volume $V$, its corresponding ground truth $G = \{g_{1}, g_{2}, ..., g_{n}\}$, $n \in \mathbb{Z}$ and obtained segmentation output $S = \{s_{1}, s_{2},...,s_{m}\}$, $m \in \mathbb{Z}$ the DSC is mathematically expressed as
\begin{equation}
    DSC\left(G, S\right) = 2~\frac{\vert G\cap S\vert}{\vert G \vert + \vert S\vert},\label{eq:dsc}
\end{equation}
where $\vert \cdot \vert$ represents the cardinality of the set. The values for DSC lay within $[0, 1]$, where the interval extremes correspond to null or exact similarity between the compared surfaces, respectively. Additionally, we consider the Wilcoxon signed-rank test to assess the statistical significance of differences among architectures. 

\subsection{Aspects to evaluate}
As we mentioned previously, this paper aims to analyse (i) overlapping patch extraction in training and testing, (ii) single and multi-modality architectures and (iii) 2D and 3D strategies. Details on these three evaluation cornerstones are discussed in the following sections.

\subsubsection{Overlapping sampling in training and testing}
\begin{figure}
    \centering
    \includegraphics[width=0.42\textwidth]{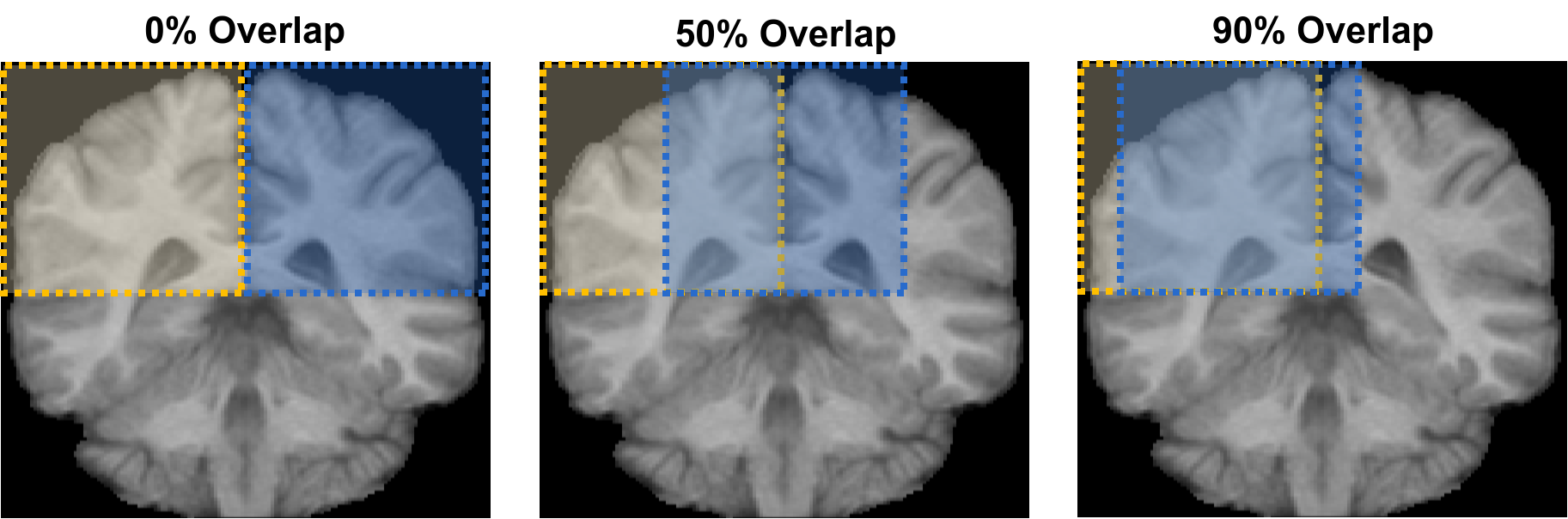}
    \caption{Patch extraction with null, medium and high overlap. Yellow and blue areas corresponds to the first and second blocks to consider. When there is overlap among patches, voxels are seen in different neighbourhoods each time.\label{fig:overlap-illustration}}
\end{figure}

\begin{figure}
    \centering
     \begin{subfigure}[b]{0.115\textwidth}
     \begin{overpic}[width=\textwidth, trim={6cm 6cm 10cm 5cm},clip]{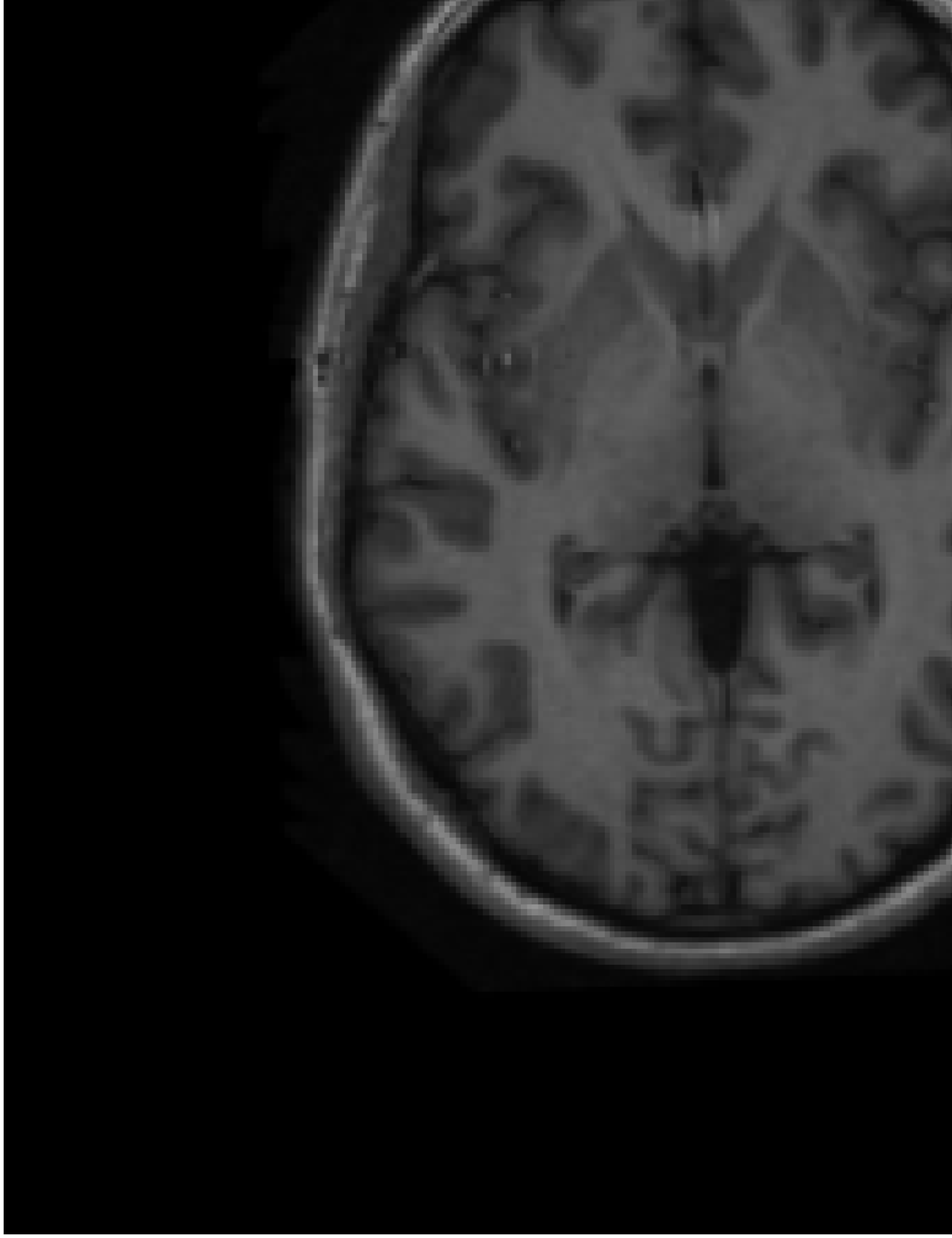}
  \linethickness{0.5mm} \color{red}%
  \put(0,0){\polygon(5,70)(5,30)(55,30)(55,70)}
 \end{overpic}
     \caption{T1-w}
     \end{subfigure}
     \begin{subfigure}[b]{0.115\textwidth}
     \begin{overpic}[width=\textwidth, trim={6cm 6cm 10cm 5cm},clip]{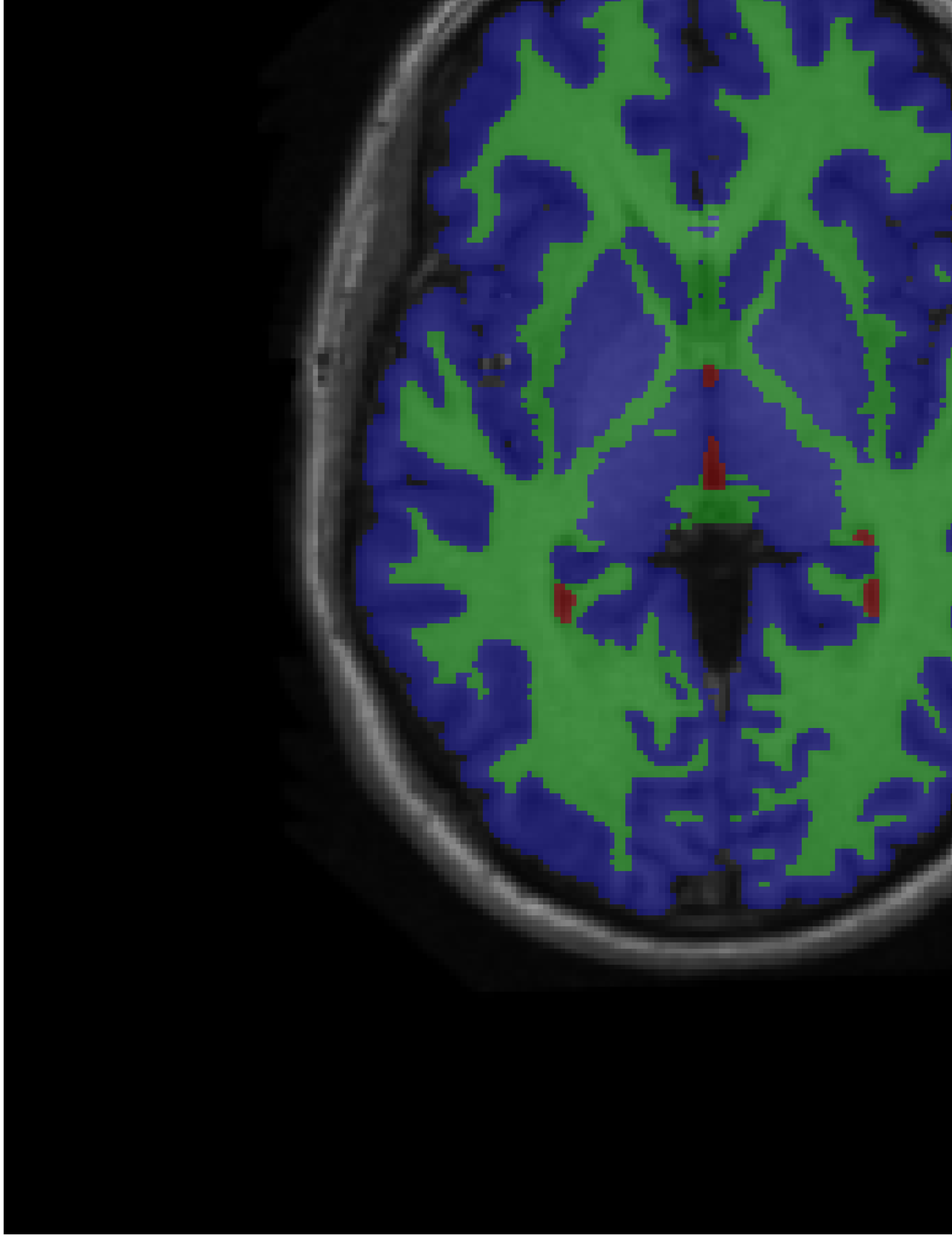}
  \linethickness{0.5mm} \color{red}%
  \put(0,0){\polygon(5,70)(5,30)(55,30)(55,70)}
 \end{overpic}
     \caption{GT}
     \end{subfigure}
     \begin{subfigure}[b]{0.115\textwidth}
     \begin{overpic}[width=\textwidth, trim={6cm 6cm 10cm 5cm},clip]{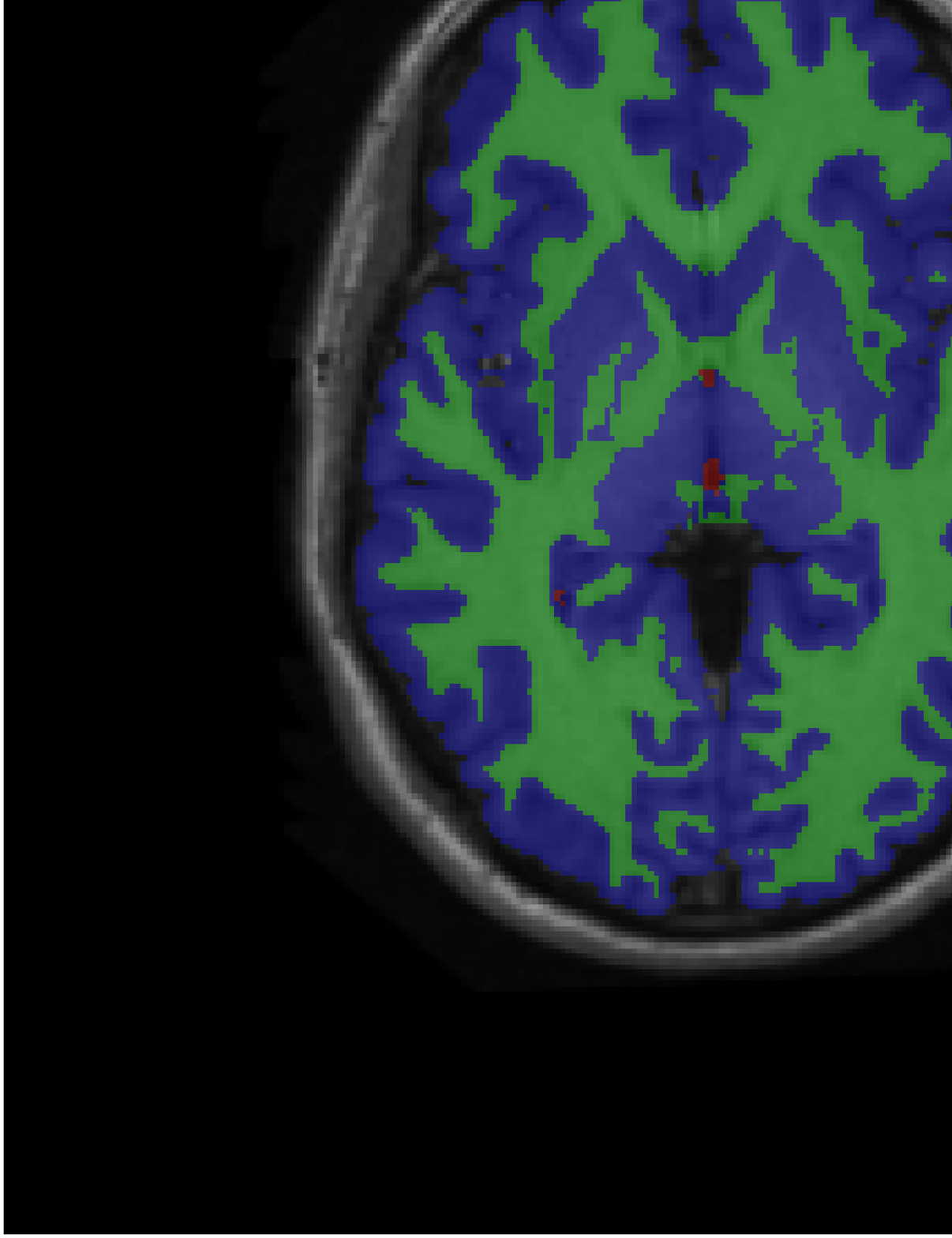}
  \linethickness{0.5mm} \color{red}%
  \put(0,0){\polygon(5,70)(5,30)(55,30)(55,70)}
 \end{overpic}
     \caption{No overlap}
     \end{subfigure}
     \begin{subfigure}[b]{0.115\textwidth}
     \begin{overpic}[width=\textwidth, trim={6cm 6cm 10cm 5cm},clip]{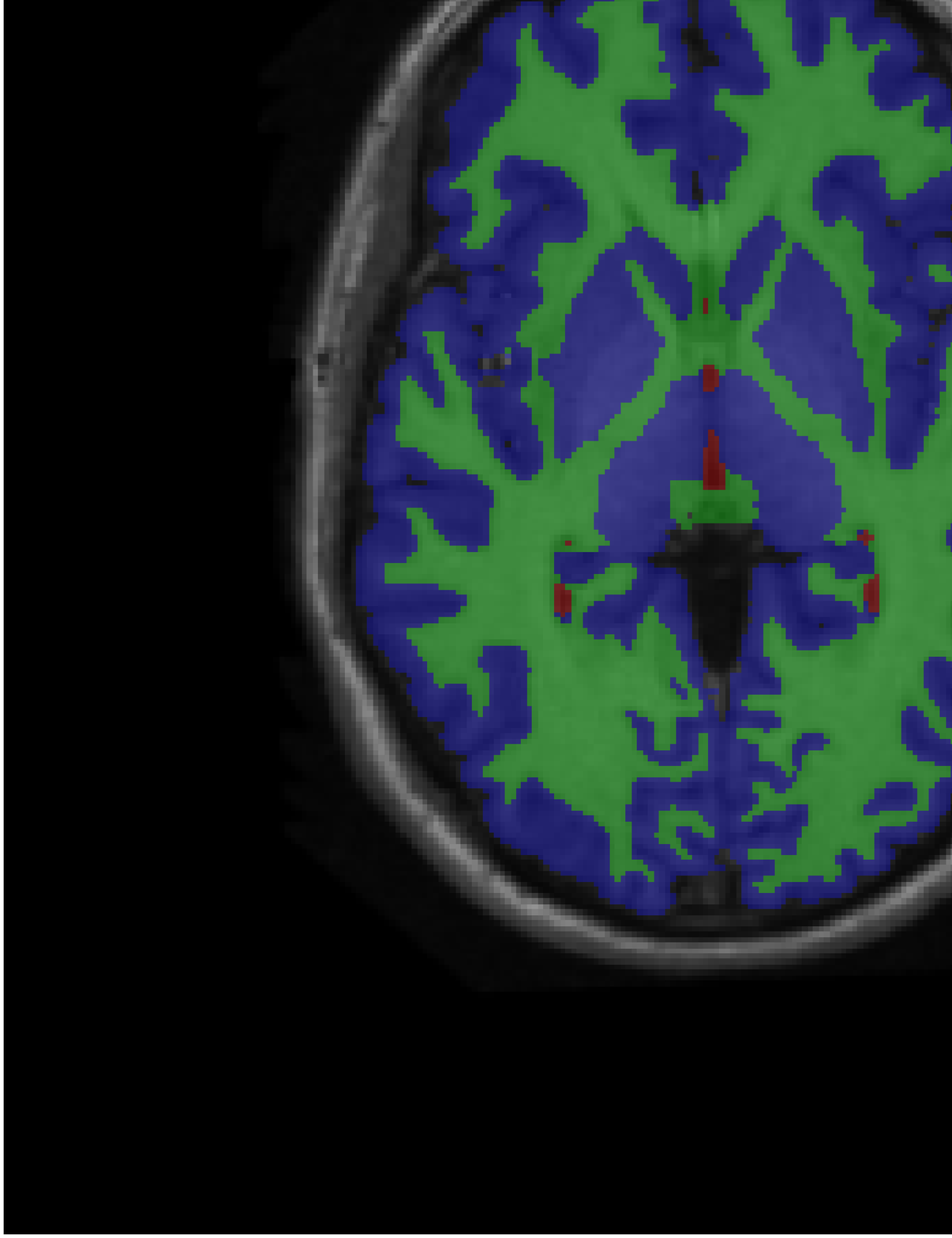}
  \linethickness{0.5mm} \color{red}%
  \put(0,0){\polygon(5,70)(5,30)(55,30)(55,70)}
 \end{overpic}
     \caption{Overlap}
     \end{subfigure}
     \begin{subfigure}[b]{0.115\textwidth}
     \begin{overpic}[width=\textwidth, trim={11cm 8cm 15cm 2.5cm},clip]{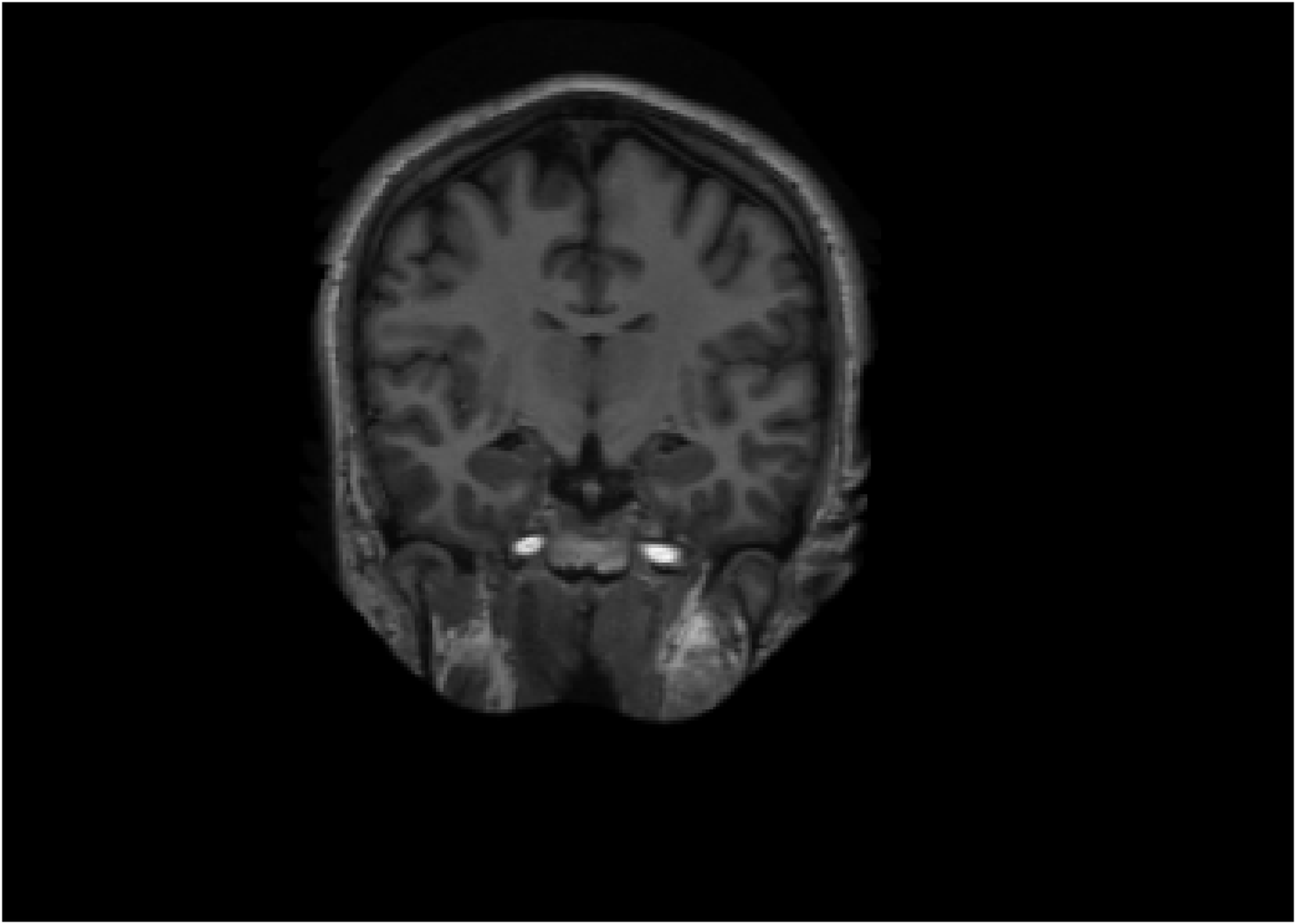}
  \linethickness{0.5mm} \color{red}%
  \put(0,0){\polygon(15,70)(15,30)(55,30)(55,70)}
 \end{overpic}
     \caption{T1-w}
     \end{subfigure}
     \begin{subfigure}[b]{0.115\textwidth}
     \begin{overpic}[width=\textwidth, trim={11cm 8cm 15cm 2.5cm},clip]{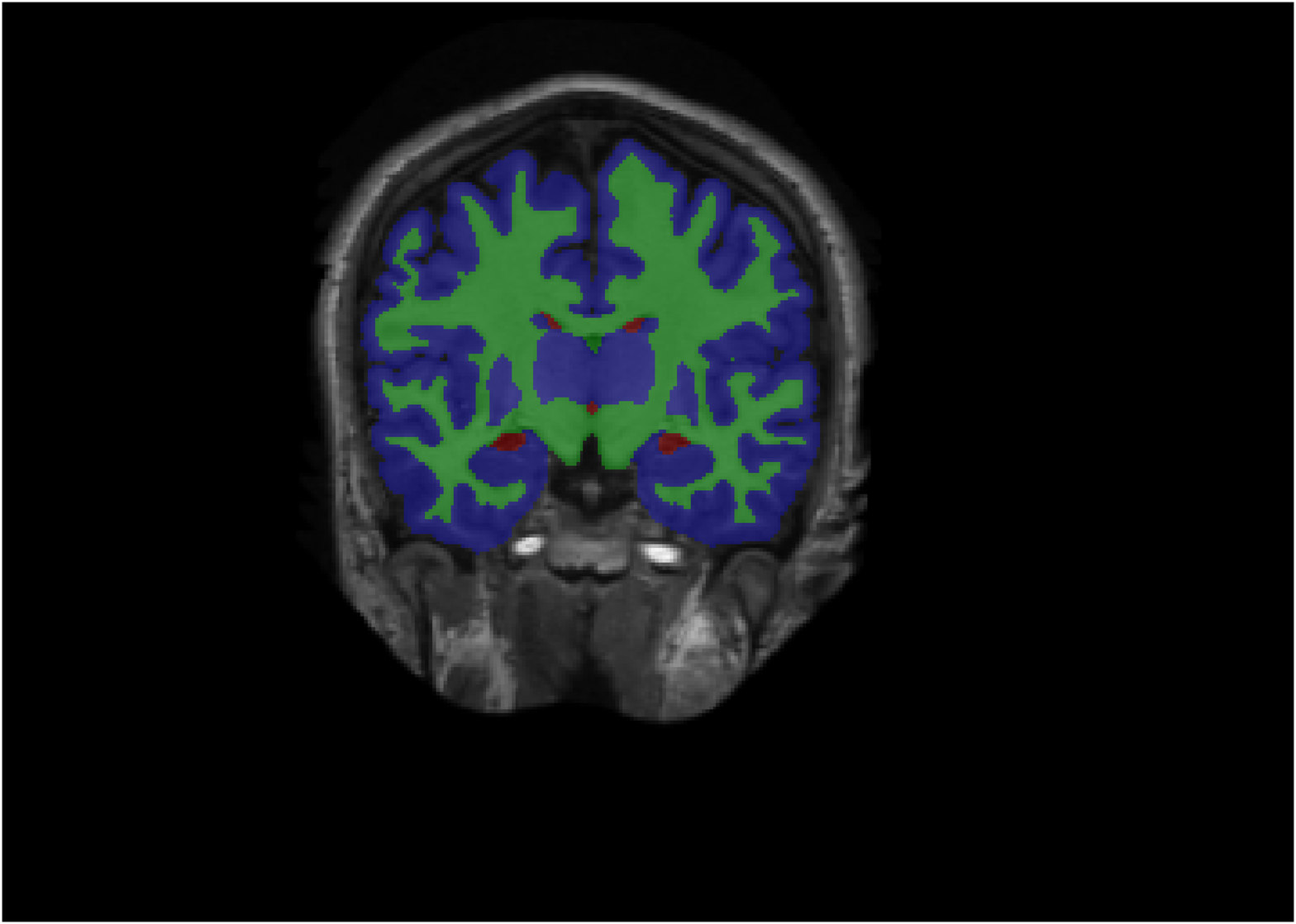}
  \linethickness{0.5mm} \color{red}%
  \put(0,0){\polygon(15,70)(15,30)(55,30)(55,70)}
 \end{overpic}
     \caption{GT}
     \end{subfigure}
     \begin{subfigure}[b]{0.115\textwidth}
     \begin{overpic}[width=\textwidth, trim={11cm 8cm 15cm 2.5cm},clip]{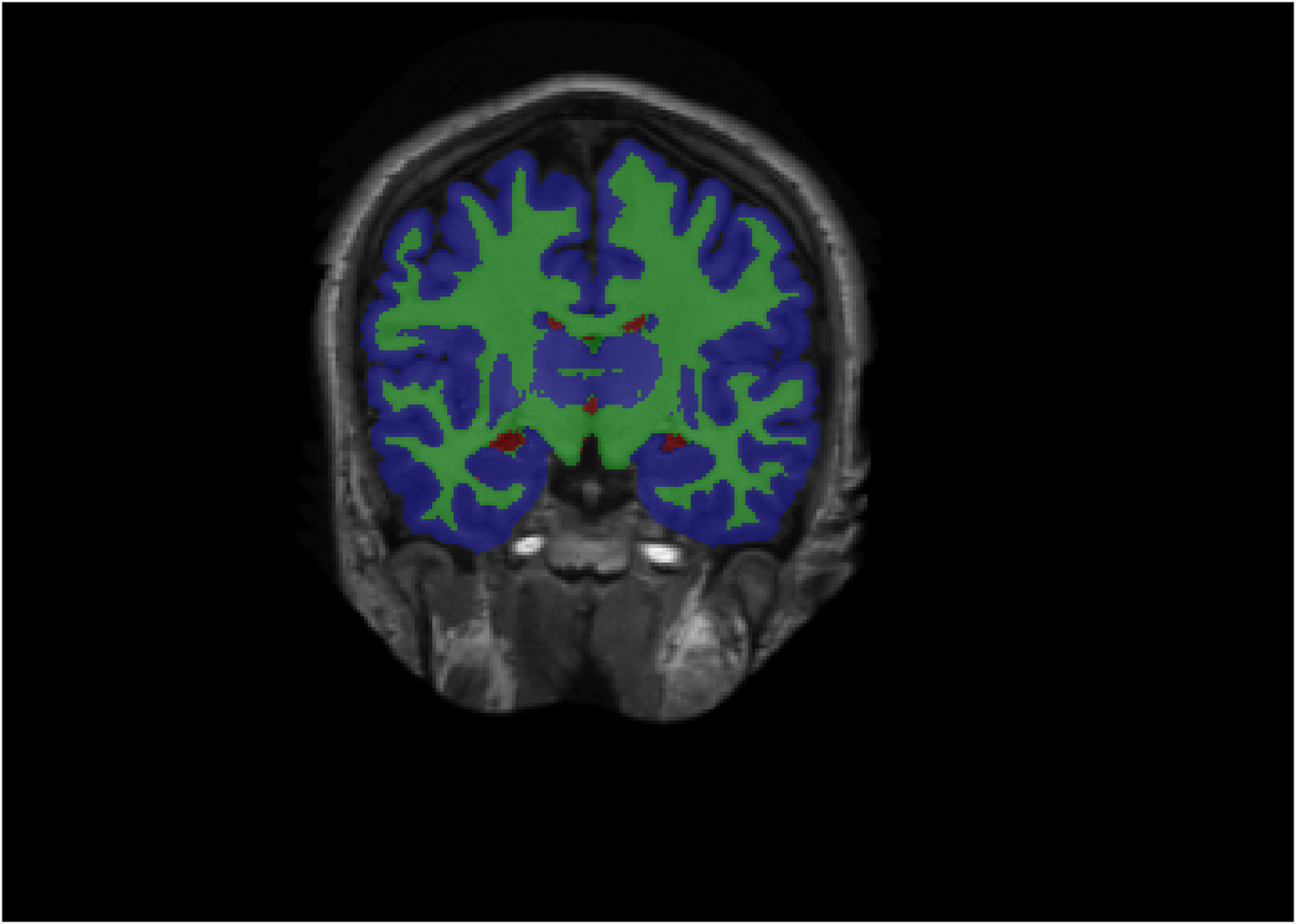}
  \linethickness{0.5mm} \color{red}%
  \put(0,0){\polygon(15,70)(15,30)(55,30)(55,70)}
 \end{overpic}
     \caption{No overlap}
     \end{subfigure}
     \begin{subfigure}[b]{0.115\textwidth}
     \begin{overpic}[width=\textwidth, trim={11cm 8cm 15cm 2.5cm},clip]{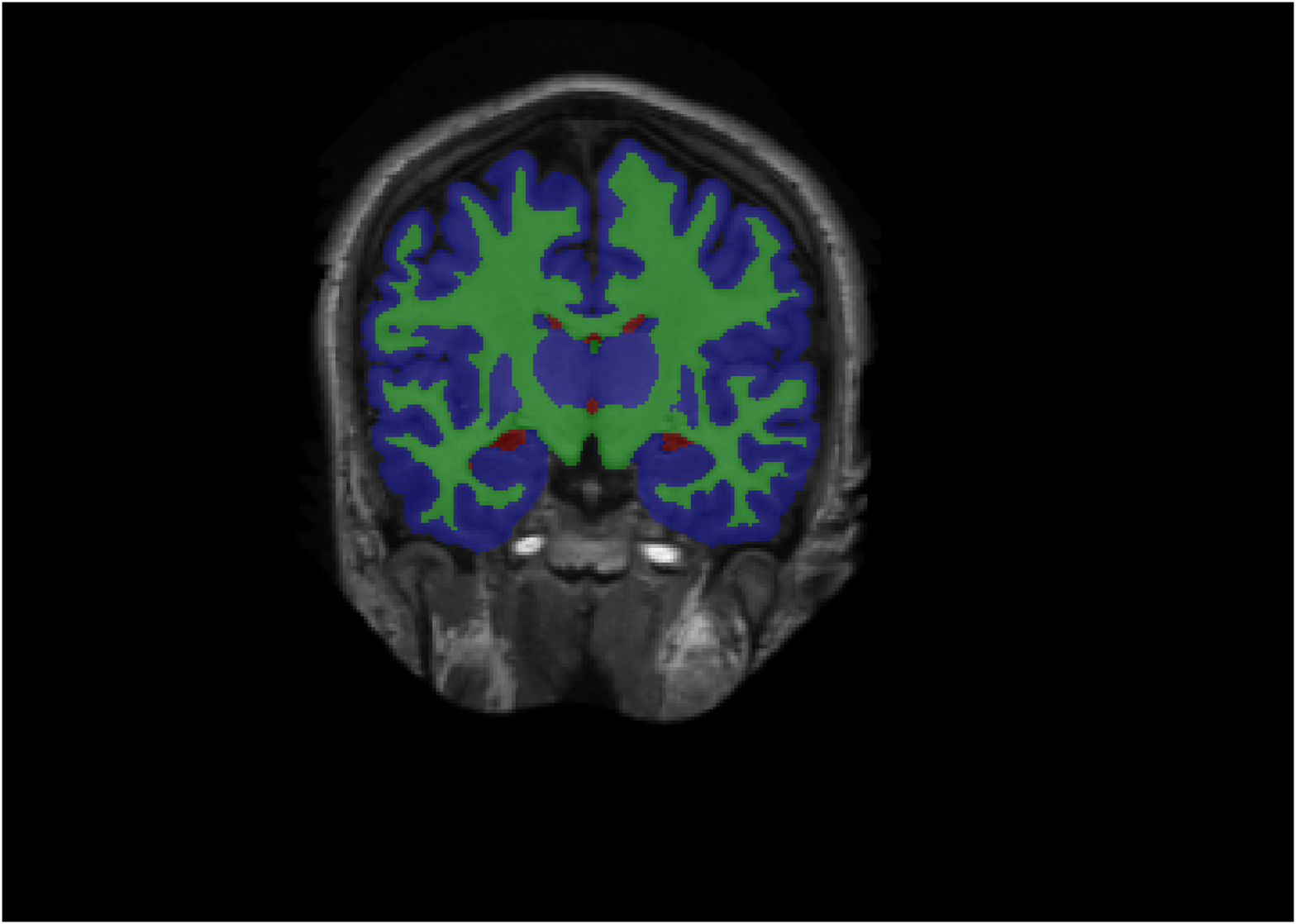}
  \linethickness{0.5mm} \color{red}%
  \put(0,0){\polygon(15,70)(15,30)(55,30)(55,70)}
 \end{overpic}
     \caption{Overlap}
     \end{subfigure}

    \caption{Segmentation using overlapping patch extraction in training (a-d) and testing (e-h). From left to right, T1-w volume (a)-(e), ground truth (b)-(f), segmentation without overlap (c)-(g) and with overlap (d)-(h). The area inside the red box depicts notable changes between strategies. Notice that results obtained with overlapping sampling appear more similar to the ground truth. Colours for CSF, GM and WM are red, blue and green, respectively. \label{fig:overlap-example}}
\end{figure}

One of the drawbacks of networks performing dense-inference is that -- under similar conditions -- the number of parameters increases. This implies that more samples should be used in training to obtain acceptable results. The most common approach consists in augmenting the input data through transformations -- e.g. translation, rotation, scaling. However, if the output dimension is not equal to the input size, other options can be considered. For instance, patches can be extracted from the input volumes with a certain extent of overlap and, thus, the same voxel would be seen several times in different neighbourhoods. An example of this technique can be observed in Fig.~\ref{fig:overlap-illustration}. Consequently, (i) more samples are gathered, and (ii) networks are provided with information that may improve spatial consistency as illustrated in Fig.~\ref{fig:overlap-example}~(a-d).

The sampling strategy aforementioned can be additionally enhanced by overlaying predicted patches. Unlike sophisticated post-processing techniques, the network itself is used to improve its segmentation. As depicted in Fig.~\ref{fig:overlap-example}~(e-h), the leading property of this post-processing technique is that small segmentation errors -- e.g. holes and block boundary artefacts -- are corrected. The consensus among outputs can be addressed through majority voting, for instance. 

\subsubsection{Input modalities}
Depending on the number of modalities available in a dataset, approaches can be either single- or multi-modality. If many modalities were acquired, networks could be adapted to process them all at the same time either using different channels or various processing paths -- also referred in the literature as early and late fusion schemes~\citep{ghafoorian2016location}, respectively. Naturally, regarding computational resources the former strategy is desirable, but the latter may extract more valuable features. In this work, we consider the early fusion only. Regardless of the fusion scheme, merging different sources of information may provide models with complementary features which may lead to enhanced segmentation results~\citep{Zhang2015}. 

\subsubsection{Network's dimensionality}
There are two streams of FCNN regarding its input dimensionality: 2D and 3D. On the one hand, 2D architectures are fast, flexible and scalable; however, they ignore completely data from neighbouring slices, i.e. implicit information is reduced compared to 3D approaches. On the other hand, 3D networks acquire valuable implicit contextual information from orthogonal planes. These strategies tend to lead to better performance than 2D -- even if labelling is carried out slice-by-slice -- but they are computationally demanding -- exponential increase in parameters and resource consumption. Moreover, due to the increasing number of parameters, these models may require larger training sets. Therefore, depending on the data itself, one approach would be more suitable than the other.

\subsection{Implementation details}
General tissue segmentation pipelines contemplate four essential components: preprocessing, data preparation, classification and post-processing. Specific implementations of each one of these elements can be plugged and unplugged as required to achieve the best performance. First, preprocessing is carried out by (i) removing skull, and (ii) normalising intensities between scans. We use the ground truth masks to address the former tasks and standardise our data to have zero mean and unit variance. Second, data is prepared by extracting useful and overlapping patches -- containing information from one of the three tissues. Third, classification takes place; segmentation of an input volume is provided through means of majority voting in case of overlapping predictions. Fourth, no post-processing technique was considered.

All the networks were trained maximum for 20 epochs. Training stopping criterium was overfitting, which was monitored using an early stopping policy with patience equal to $2$. For each of the datasets, we split the training set into training and validation (80\% and 20\% of the volumes, respectively). Additionally, voxels laying on the background area were given a weight of zero to avoid considering them in the optimisation/training process.

All the architectures have been implemented from scratch in Python, using the Keras library. From here on, our implementations of \cite{Dolz2017}, \cite{kamnitsas2017efficient}, \cite{cciccek20163d}, and \cite{guerrero2017white} are denoted by $DM$, $KK$, $UNet$ and $UResNet$, respectively. The number of parameters per architecture is listed in Table~\ref{tab:parameters}. All the experiments have been run on a GNU/Linux machine box running Ubuntu 16.04, with 128GB RAM. CNN training and testing have been carried out using a single TITAN-X PASCAL GPU (NVIDIA corp., United States) with $8$GB RAM. The developed framework for this work is currently available to download at our research website. The source code includes architecture implementation and experimental evaluation scripts.

\begin{table}
    \centering
    {\footnotesize
    \caption{Number of parameters per considered architecture and per dimensionality.\label{tab:parameters}}
    \begin{tabular}{|c|c|c|c|c|}
    \hline
    \textbf{Dimensionality} & \textbf{DM} & \textbf{KK} & \textbf{UNet} & \textbf{UResNet} \\ \hline
    2D & 569,138 & 547,053 & 1,930,756 & 994,212 \\ \hline
    3D & 7,099,418 & 3,332,595 & 5,605,444 & 2,622,948 \\ \hline
    \end{tabular}}
\end{table}

\section{Results\label{sec:results}}
The evaluation conducted in this paper is three-fold. First, we investigate the effect of overlapping patches in both training and testing stages. Second, we assess the improvement of multi-modality architectures over single-modality ones. Third, we compare the different models on the three considered datasets. Note that, for the sake of simplicity, the network' dimensionality is shown as subscript (e.g. UResNet$_{2D}$ denotes the 2D version of the UResNet architecture).

\subsection{Overlapping}
To evaluate the effect of extracting overlapping patches in training and testing, we run all the architectures on the three datasets contemplating three levels: null, medium and high (approximately 0\%, 50\% and 90\%, respectively). On IBSR18 and iSeg2017, the evaluation was carried out using a leave-one-out cross-validation scheme. On MICCAI2012, the process consisted in using the given training and testing sets. 

\begin{figure}
    \centering
    \includegraphics[width=0.35\textwidth]{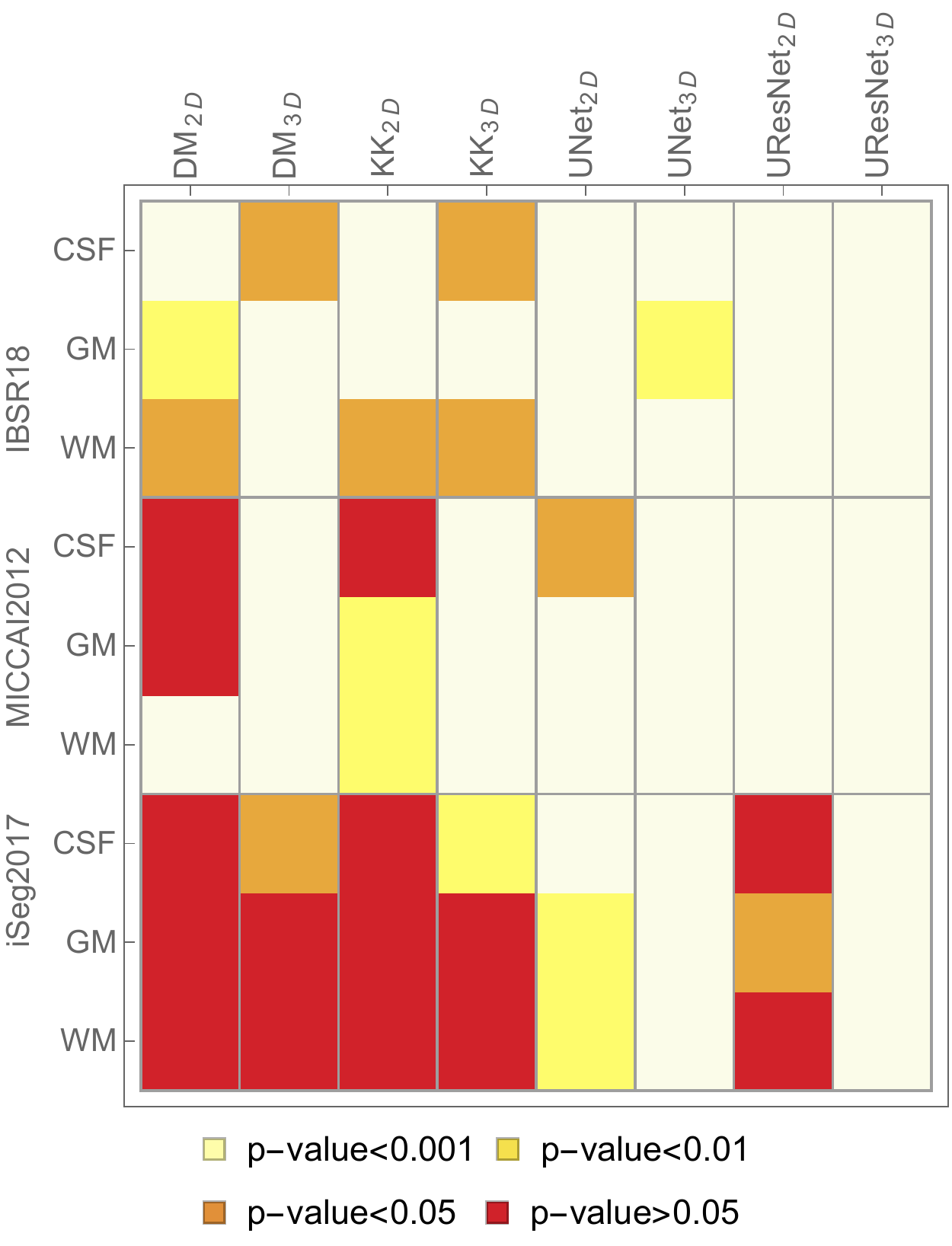}
    \caption{$p$-values obtained when comparing DSC values when using null and high overlapping as sampling strategy only in training. From top to bottom, CSF, GM and WM values for IBSR18, MICCAI2012 and iSeg2017. In all the cases the model trained with high overlap led to higher performance than when not.\label{fig:results-overlap}}
\end{figure}

The first test consisted in quantifying improvement between networks trained with either null or high degrees of overlap on training. Resulting $p$-values obtained on the three datasets are depicted in Fig.~\ref{fig:results-overlap}. In all the cases, the model trained with high overlap led to higher DSC values than when not. As it can be observed, in most of the cases ($49$ out of $72$), the overlapping sampling led to statistically significant higher performance than when omitted. The two groups (convolutional-only and u-shaped) exhibited opposite behaviours. On the one hand, the highest improvements are noted in u-shaped networks. This is related to the fact that non-overlap may mean not enough samples. On the other hand, convolutional-only models evidenced the least increase. Since output patches are smaller, more data can be extracted and passed during training. Therefore, they can provide already accurate results. This fact is illustrated by the results of DM$_{2D}$ and KK$_{2D}$.

The second test consisted in quantifying the improvement of extracting patches using combinations of the three considered degrees of overlap during training and testing. As mentioned previously, results were fused using a majority voting technique. We noted that the general trend is that the difference between results using null and high extends of overlap on prediction is not significant ($p$-values $>0.05$). Also, IQR remained the same regardless of the method or dataset. Nevertheless, the general trend was an improvement of mean DSC of at least $1$\% in the overlapping cases. Another important observation from our experiments is that zero impact or slight degradation of the DSC values was noted when training with null overlap and testing with high overlap. Naturally, this last outcome is a consequence of merging predictions of a poorly trained classifier.

Medium level of overlap patch extraction, in both training and testing, led to improvement w.r.t. null degree cases but yielded lower values than when using a high extent of overlap. That is to say, the general trend is: the more the extent of overlap, the higher the overall performance of the method. The price to pay for using farther levels of overlap is computational time and power since the number of samples to process increases proportionally. For example, given an input volume with dimensions $256\times 256\times 256$ and a network producing output size of $32\times 32\times 32$, the number of possible patches to be extracted following the null, medium and high overlap policies are $512$, $3375$ and $185193$, respectively. 

Since overlapping sampling proved useful, the results showed in following sections correspond to the ones obtained using high overlap in both training and testing. 

\subsection{Single and multiple modalities}
\begin{figure}
    \centering
    \includegraphics[width=0.45\textwidth]{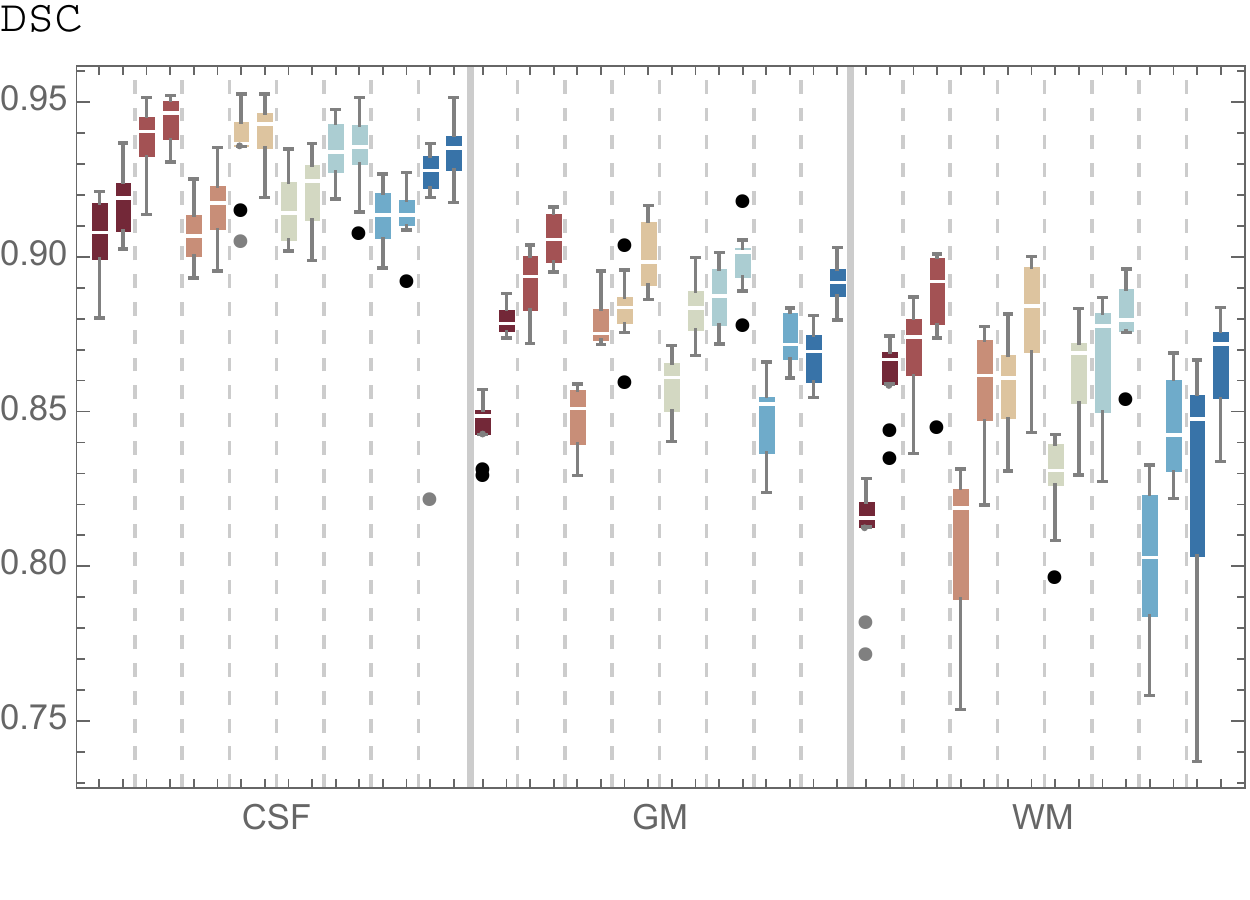}\\\vspace{-0.3cm}
    \includegraphics[width=0.45\textwidth]{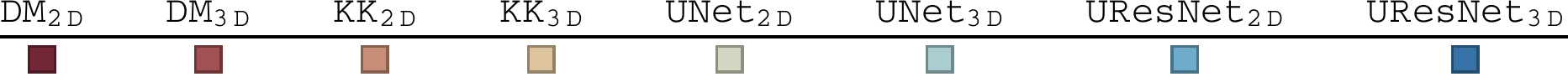}
    \caption{Evaluating the impact of single or multiple modalities for tissue segmentation on the iSeg2017 dataset. The DSC values displayed in the plot were obtained through leave-one-out cross-validation. In the plot, the same colour is used to represent each pair of single- and multi-modality versions of the same architecture. For each pair, left and right indicate whether the model considers a unique sequence or various, respectively. In the legend, subscripts indicate the dimensionality of the architecture. According to our experiments, all the multi-modality architectures outperformed significantly their single-modality analogue for GM and WM.\label{fig:modalities}}
\end{figure}

We performed leave-one-out cross-validation on the iSeg2017 dataset using the implemented 2D and 3D architectures to assess the effect of single and numerous imaging sequences on the final segmentation. The results of this experiment are shown in Fig.~\ref{fig:modalities}. 

As it can be observed, the more the input modalities, the more improved the segmentation. In this case, two modalities not only allowed the network to achieve higher mean but also to reduce the IQR, i.e. results are consistently better. This behaviour was evidenced regardless of architectural design or tissue type. For instance, while the best single modality strategy scored $0.937 \pm 0.011$, $0.891 \pm 0.010$ and $0.868 \pm 0.016$ for CSF, GM and WM, respectively; its multi-modality analogue yielded $0.944 \pm 0.008$, $0.906 \pm 0.008$ and $0.887 \pm 0.017$ for the same classes. Additionally, most of the strategies using both T1-w and T2-w obtained DSC values which were statistically higher than their single-modality counterparts ($16$ out of the $8 \times 3$ cases returned $p$-values $<0.01$). 

\subsection{Comparison of 2D and 3D FCNN architectures}
The eight architectures were evaluated using their best parameters according to the previous sections on three different datasets: IBSR18, MICCAI2012 and iSeg2017. The DSC mean and standard deviation values are shown in Table~\ref{tab:comparison}.

\begin{table*}[t]
    \centering
    \caption{DSC values obtained by DM, KK, UResNet and UNet on three analysed datasets. The highest DSC values per class and per dataset are in bold. The values with an asterisk (*) indicate that a certain architecture obtained a significantly higher score ($p$-value $<0.01$) than its analogue.\label{tab:comparison}}
    {\footnotesize
    \resizebox{1 \textwidth}{!}{
    \begin{tabular}{c|c|c|c|c|c|c|c|c|c|}
    \cline{2-10}
    & \multirow{2}{*}{\textbf{Class}} & \multicolumn{2}{c|}{\textbf{DM}} & \multicolumn{2}{c|}{\textbf{KK}} & \multicolumn{2}{c|}{\textbf{UResNet}} & \multicolumn{2}{c|}{\textbf{UNet}} \\ \cline{3-10}
    & & $\mathbf{2D}$ & $\mathbf{3D}$ & $\mathbf{2D}$ & $\mathbf{3D}$ & $\mathbf{2D}$ & $\mathbf{3D}$ & $\mathbf{2D}$ & $\mathbf{3D}$ \\
    \hline
    \hline
    \parbox[t]{2mm}{\multirow{3}{*}{\rotatebox[origin=c]{90}{{\scriptsize IBSR18}}}} & CSF & $0.87 \pm 0.05$ & $0.86 \pm 0.07$ & $0.88 \pm 0.04^*$ & $0.80 \pm 0.20$ & $\mathbf{0.90 \pm 0.03}$ & $0.89 \pm 0.05$ & $\mathbf{0.90 \pm 0.03}$ & $0.88 \pm 0.05$ \\ \cline{2-10}
    & GM & $\mathbf{0.96 \pm 0.01}$ & $\mathbf{0.96 \pm 0.01}$ & $0.88 \pm 0.04$ & $\mathbf{{0.96 \pm 0.01}^*}$ & $\mathbf{0.96 \pm 0.01}$ & $\mathbf{0.96 \pm 0.01}$ & $\mathbf{0.96 \pm 0.01}$ & $\mathbf{0.96 \pm 0.01}$ \\ \cline{2-10}
    & WM & $0.92 \pm 0.02$ & $\mathbf{0.93 \pm 0.02^*}$ & $0.92 \pm 0.02$ & $0.92 \pm 0.02$ & $\mathbf{0.93 \pm 0.02}$ & $\mathbf{0.93 \pm 0.02}$ & $\mathbf{0.93 \pm 0.02}$ & $\mathbf{0.93 \pm 0.02}$ \\ \hline \hline

    \parbox[t]{2mm}{\multirow{3}{*}{\rotatebox[origin=c]{90}{{\scriptsize MICCAI}}}} & CSF & $0.87 \pm 0.04$ & $\mathbf{0.91 \pm 0.02^*}$ & $0.81 \pm 0.10$ & $0.90 \pm 0.03^*$ & $0.89 \pm 0.04$ & $0.91 \pm 0.03^*$ & $0.90 \pm 0.04$ & $0.91 \pm 0.03^*$ \\ \cline{2-10}
    & GM & $0.95 \pm 0.02$ & $0.96 \pm 0.01^*$ & $0.95 \pm 0.02$ & $0.96 \pm 0.01^*$ & $0.96 \pm 0.02$ & $0.96 \pm 0.02^*$ & $0.96 \pm 0.02$ & $\mathbf{0.97 \pm 0.01^*}$ \\ \cline{2-10}
    & WM & $0.92 \pm 0.03$ & $0.94 \pm 0.02^*$ & $0.93 \pm 0.02$ & $0.94 \pm 0.02^*$ & $0.93 \pm 0.03$ & $0.94 \pm 0.02^*$ & $0.94 \pm 0.03$ & $\mathbf{0.95 \pm 0.02^*}$ \\ \hline \hline
    
    \parbox[t]{2mm}{\multirow{3}{*}{\rotatebox[origin=c]{90}{{\scriptsize iSeg2017}}}} & CSF & $0.92 \pm 0.01$ & $\mathbf{0.94 \pm 0.01^*}$ & $0.92 \pm 0.01$ & $\mathbf{0.94 \pm 0.01^*}$ & $0.91 \pm 0.01$ & $0.93 \pm 0.01^*$ & $0.92 \pm 0.01$ & $0.93 \pm 0.01^*$ \\ \cline{2-10}
    & GM & $0.88 \pm 0.01$ & $\mathbf{0.91 \pm 0.01^*}$ & $0.88 \pm 0.01$ & $0.90 \pm 0.01^*$ & $0.87 \pm 0.01$ & $0.89 \pm 0.01^*$ & $0.88 \pm 0.01$ & $0.90 \pm 0.01^*$ \\ \cline{2-10}
    & WM & $0.86 \pm 0.01$ & $\mathbf{0.89 \pm 0.02^*}$ & $0.86 \pm 0.02$ & $0.88 \pm 0.02^*$ & $0.85 \pm 0.02$ & $0.87 \pm 0.02^*$ & $0.86 \pm 0.02$ & $0.88 \pm 0.01^*$ \\ \hline

    \hline
    \end{tabular}}}
\end{table*}

The networks performing the best on IBSR18, MICCAI2012 and iSeg2017 were UResNet$_{2D}$ and UNet$_{2D}$, UNet$_{3D}$, and DM$_{3D}$. While for MICCAI2012 and iSeg2017, 3D approaches took the lead, 2D versions performed the best for IBSR18. Taking into account the information in Table~\ref{tab:dataset-information}, 3D architectures appear to be slightly more affected by differences in voxel spacing than 2D ones since the former set obtains similar or lower results than latter group -- unlike in the other datasets. One of the reasons explaining this outcome could be the lack of sufficient data allowing 3D networks to understand variations in voxel spacing, i.e. 3D networks may be overfitting to one of the voxel spacing groups. Nevertheless, the increase of the 2D networks w.r.t. the 3D ones is not statistically significant overall. 

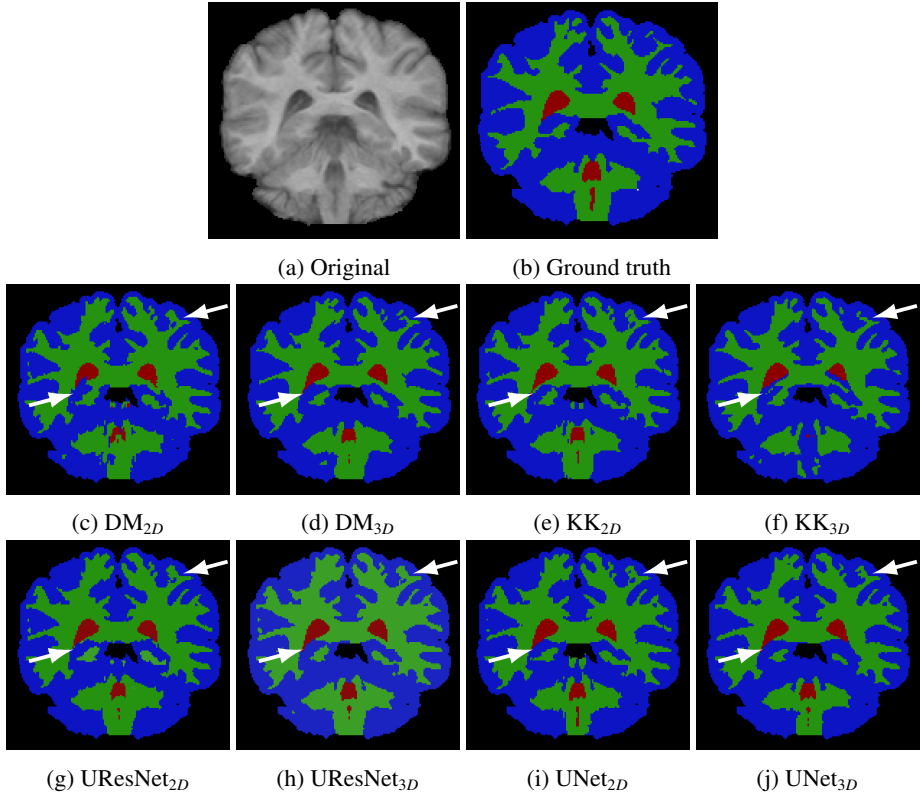
\begin{figure*}
    \centering
     \begin{subfigure}[b]{0.18\textwidth}
     \begin{overpic}[width=\textwidth, trim={7cm 8cm 7cm 7cm},clip]{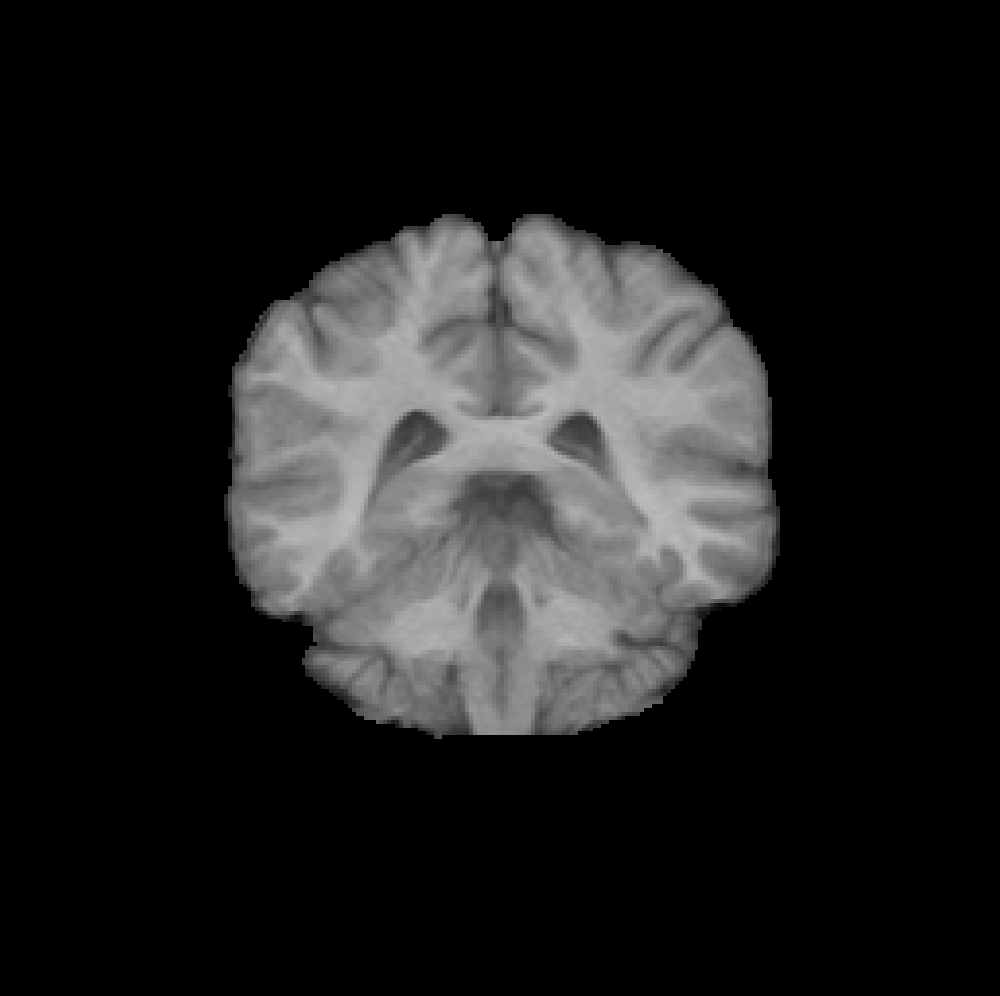}
     \linethickness{0.5mm} \color{red}%
     \end{overpic}
     \caption{Original}
     \end{subfigure}
     \begin{subfigure}[b]{0.18\textwidth}
     \begin{overpic}[width=\textwidth, trim={7cm 8cm 7cm 7cm},clip]{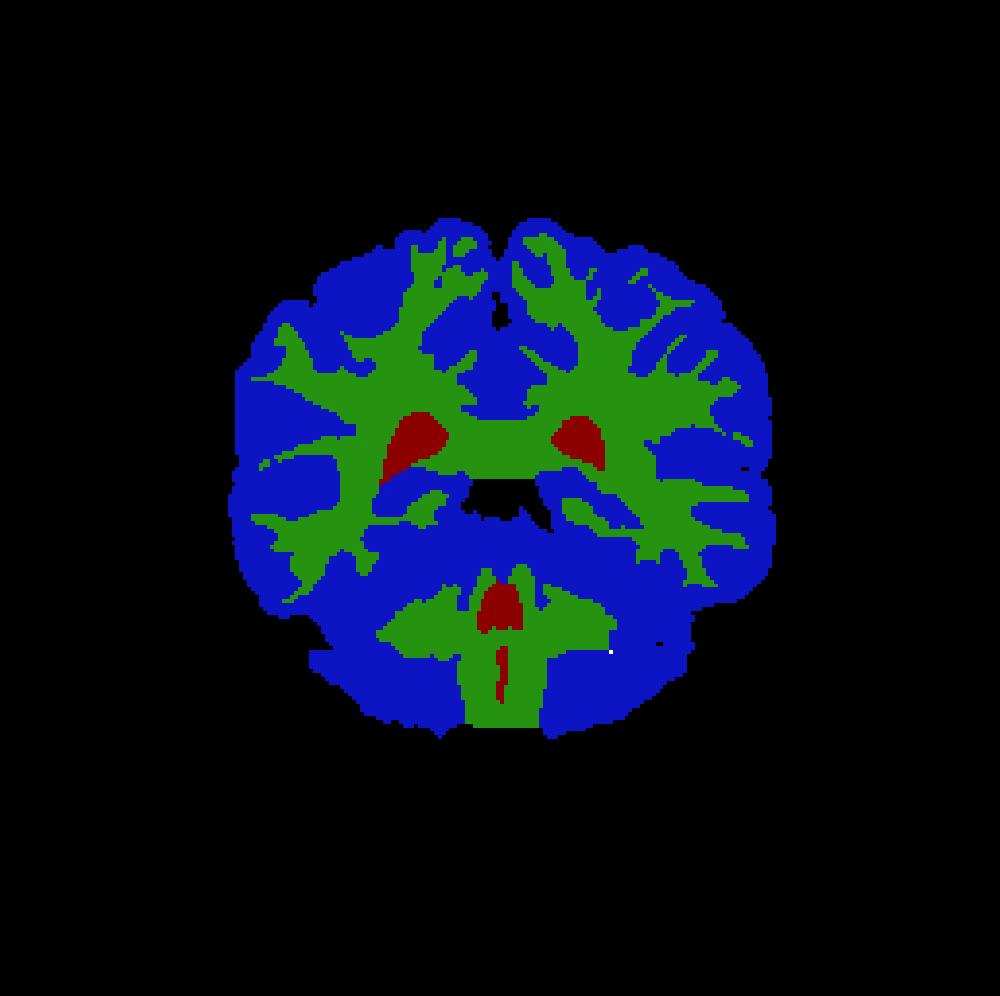}
     \linethickness{0.5mm} \color{red}%
     \end{overpic}
     \caption{Ground truth}
     \end{subfigure}
     \\
     \begin{subfigure}[b]{0.16\textwidth}
     \begin{overpic}[width=\textwidth, trim={7cm 8cm 7cm 7cm},clip]{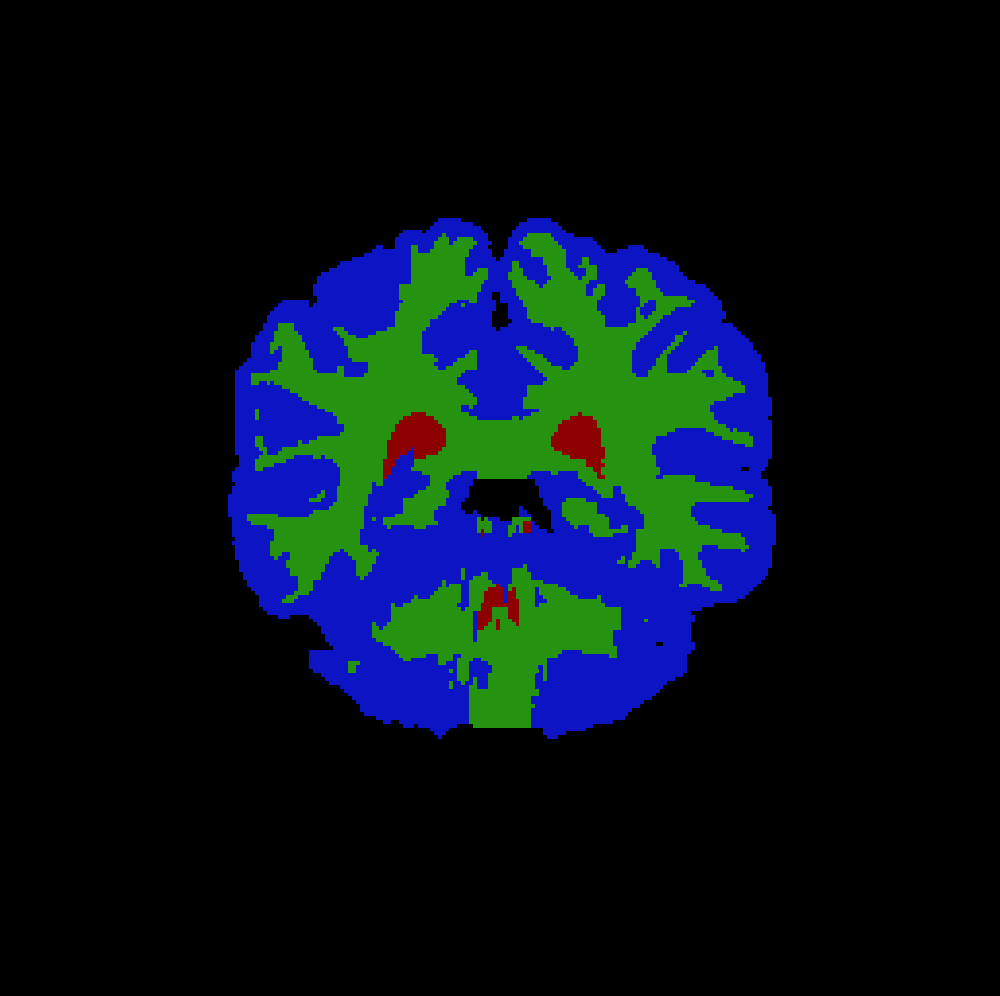}
     \linethickness{0.5mm} \color{white}%
      \put(90,85){\vector(-966,-259){20}}
     \put(1,40){\vector(966,259){20}}
     \end{overpic}
     \caption{DM$_{2D}$}
     \end{subfigure}
     \begin{subfigure}[b]{0.16\textwidth}
     \begin{overpic}[width=\textwidth, trim={7cm 8cm 7cm 7cm},clip]{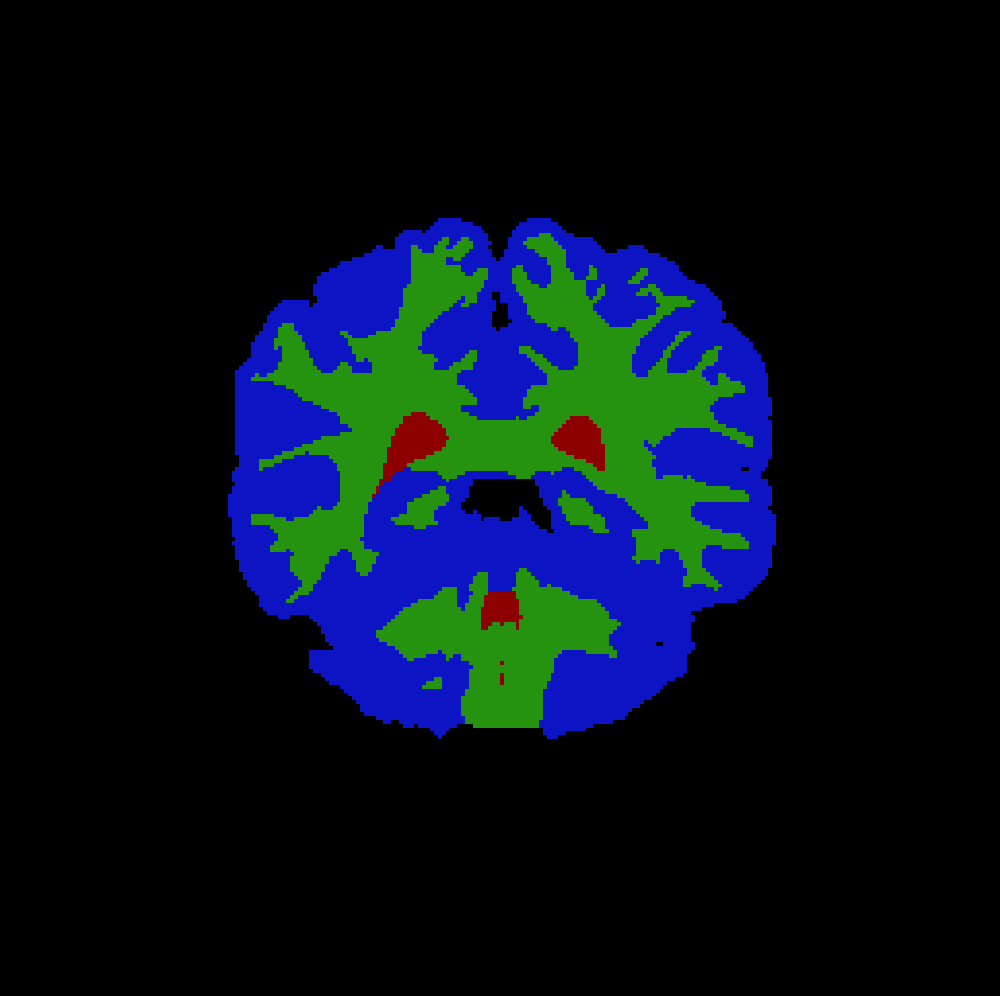}
     \linethickness{0.5mm} \color{white}%
      \put(90,85){\vector(-966,-259){20}}
     \put(1,40){\vector(966,259){20}}
     \end{overpic}
     \caption{DM$_{3D}$}
     \end{subfigure}
     \begin{subfigure}[b]{0.16\textwidth}
     \begin{overpic}[width=\textwidth, trim={7cm 8cm 7cm 7cm},clip]{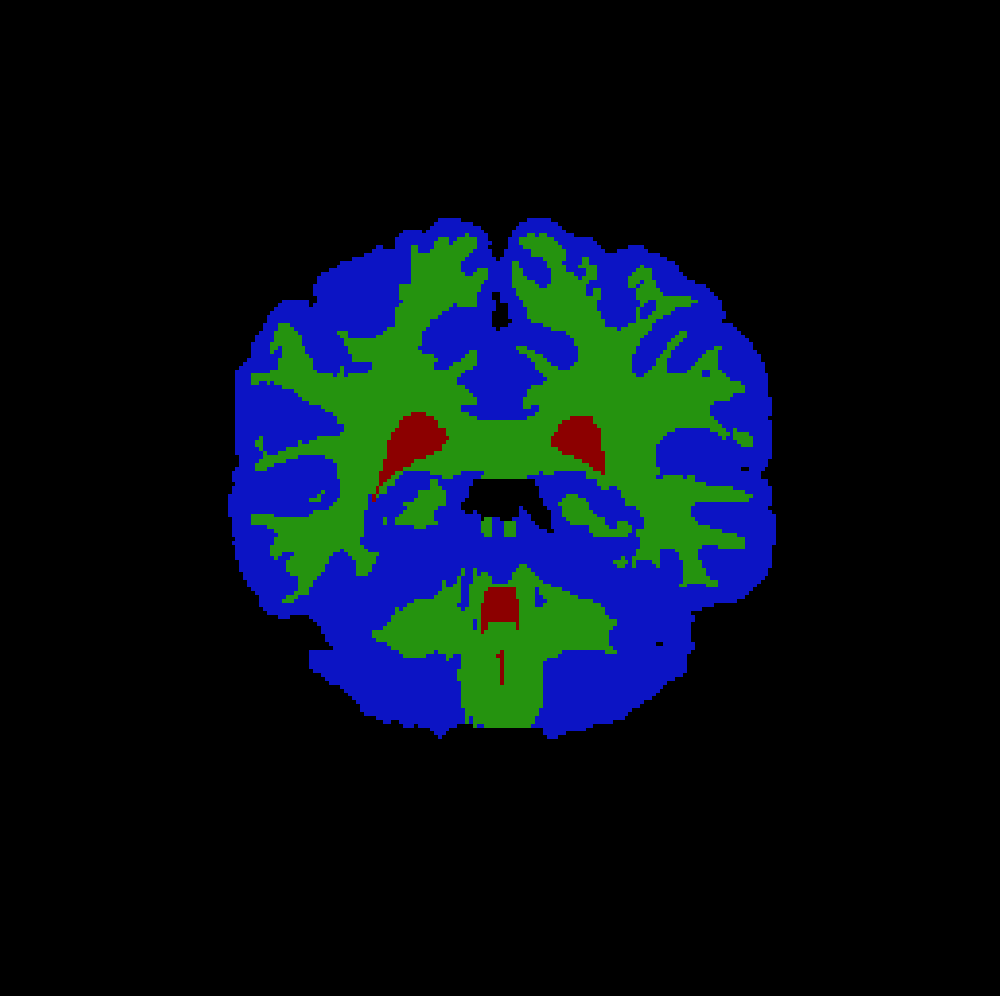}
     \linethickness{0.5mm} \color{white}%
      \put(90,85){\vector(-966,-259){20}}
     \put(1,40){\vector(966,259){20}}
     \end{overpic}
     \caption{KK$_{2D}$}
     \end{subfigure}
     \begin{subfigure}[b]{0.16\textwidth}
     \begin{overpic}[width=\textwidth, trim={7cm 8cm 7cm 7cm},clip]{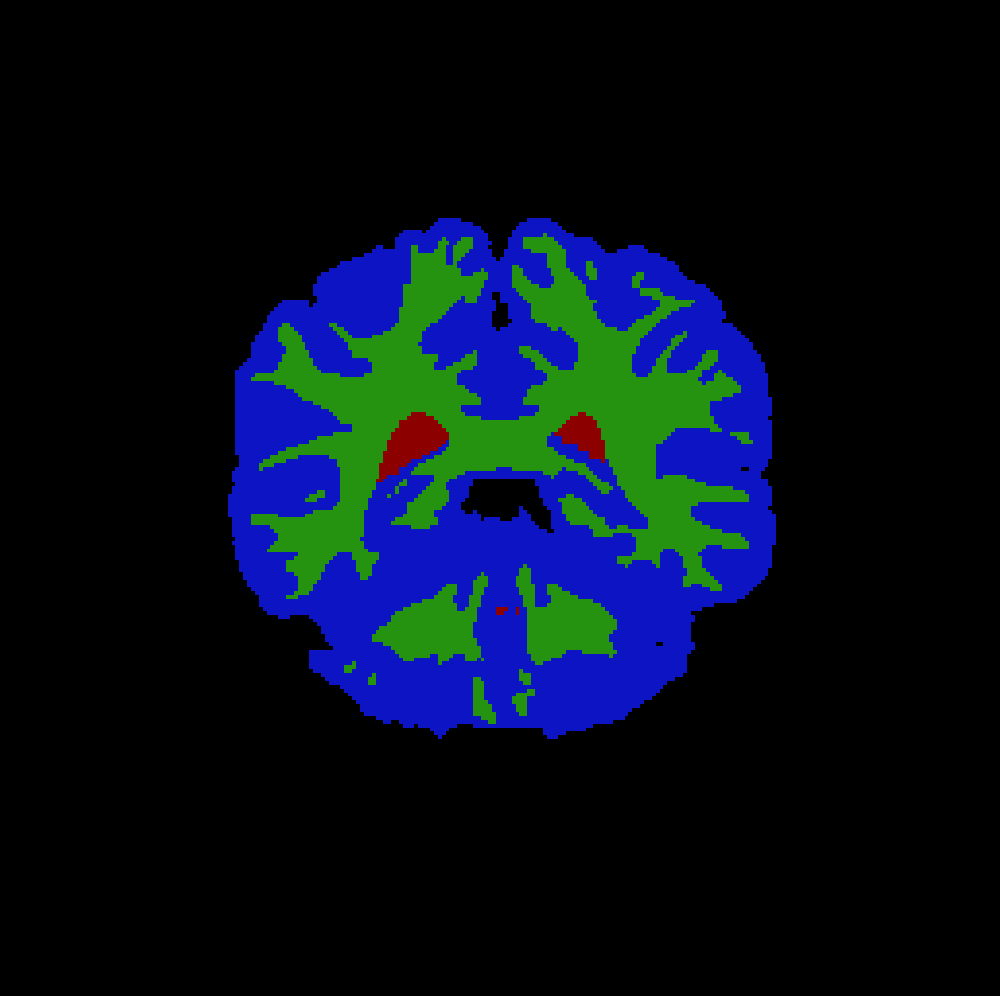}
     \linethickness{0.5mm} \color{white}%
      \put(90,85){\vector(-966,-259){20}}
     \put(1,40){\vector(966,259){20}}
     \end{overpic}
     \caption{KK$_{3D}$}
     \end{subfigure}\\
     \begin{subfigure}[b]{0.16\textwidth}
     \begin{overpic}[width=\textwidth, trim={7cm 8cm 7cm 7cm},clip]{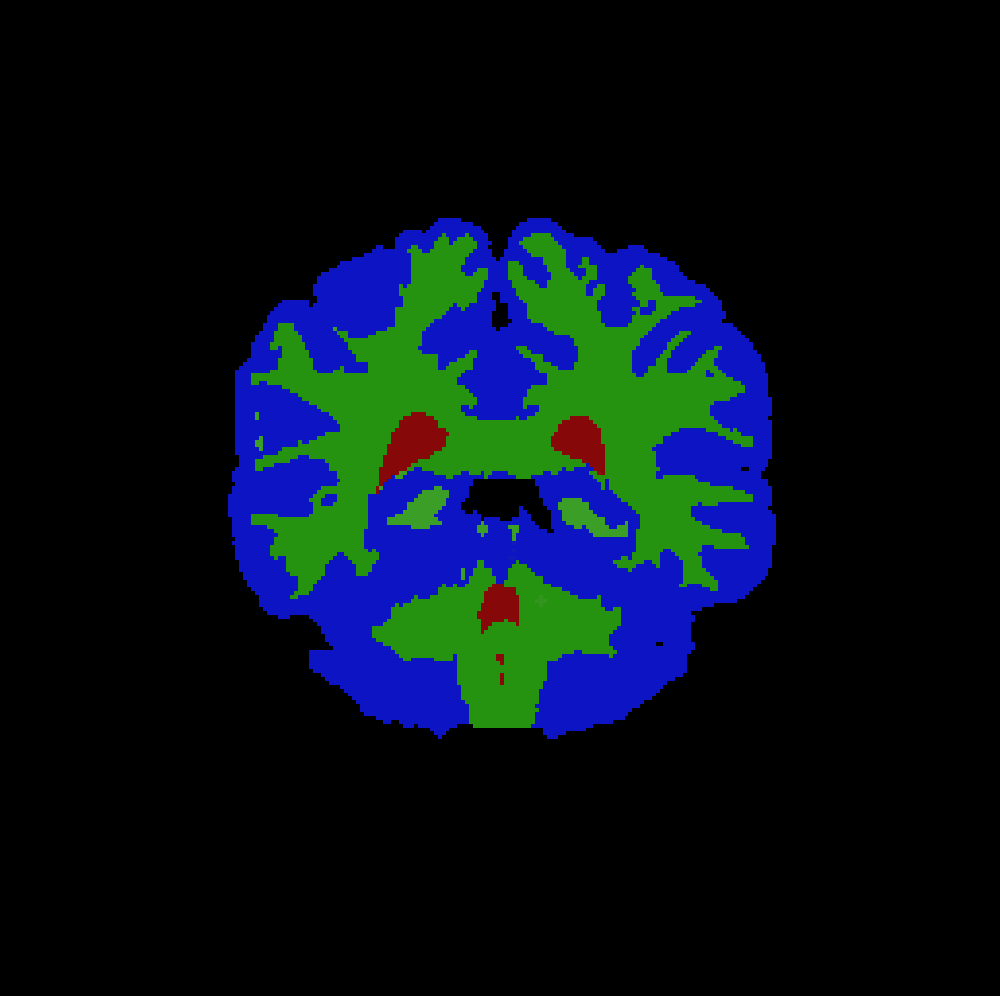}
     \linethickness{0.5mm} \color{white}%
      \put(90,85){\vector(-966,-259){20}}
     \put(1,40){\vector(966,259){20}}
     \end{overpic}
     \caption{UResNet$_{2D}$}
     \end{subfigure}
     \begin{subfigure}[b]{0.16\textwidth}
     \begin{overpic}[width=\textwidth, trim={7cm 8cm 7cm 7cm},clip]{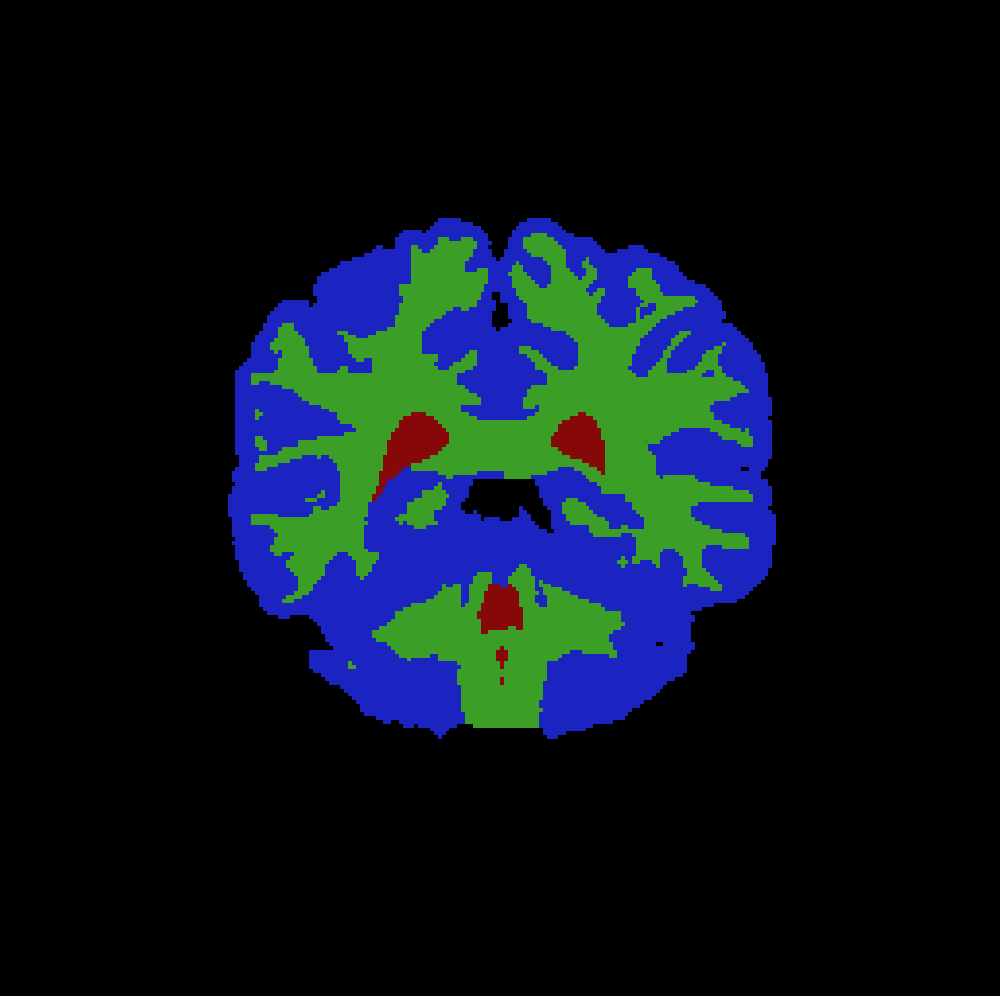}
     \linethickness{0.5mm} \color{white}%
      \put(90,85){\vector(-966,-259){20}}
     \put(1,40){\vector(966,259){20}}
     \end{overpic}
     \caption{UResNet$_{3D}$}
     \end{subfigure}
     \begin{subfigure}[b]{0.16\textwidth}
     \begin{overpic}[width=\textwidth, trim={7cm 8cm 7cm 7cm},clip]{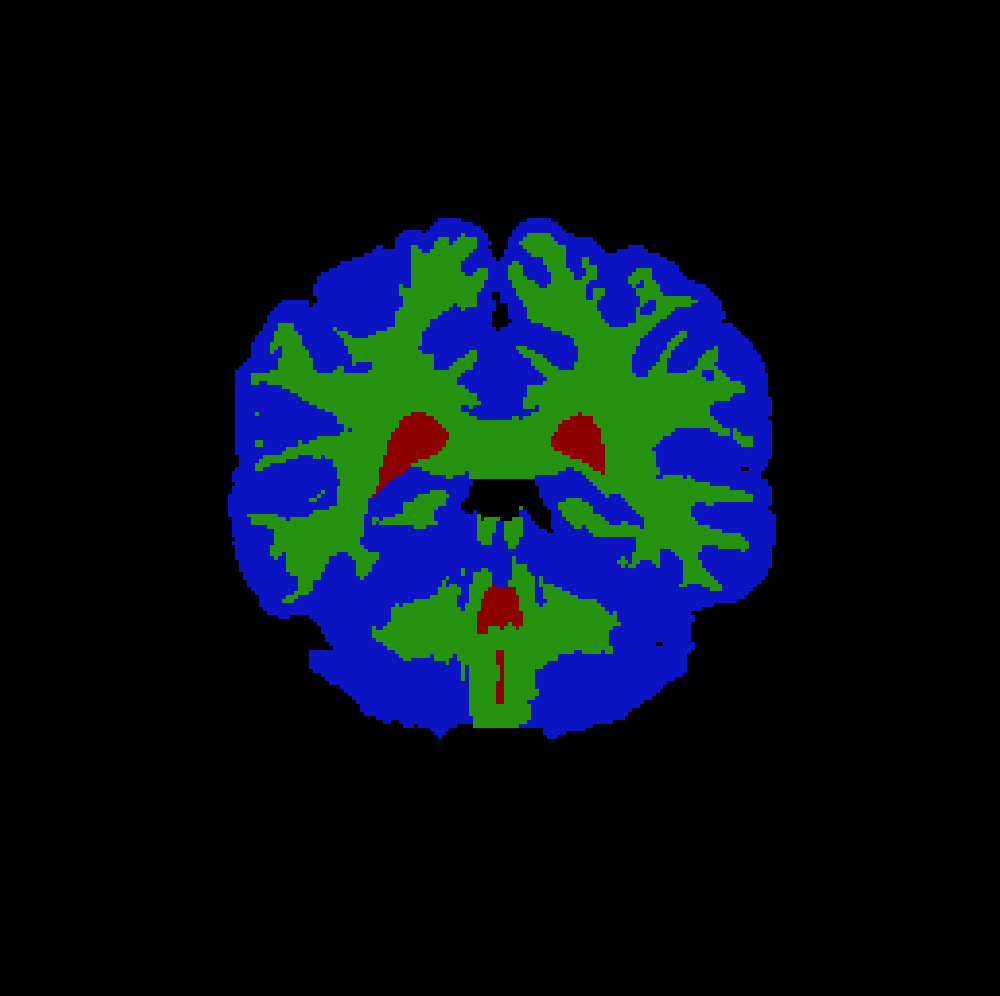}
     \linethickness{0.5mm} \color{white}%
      \put(90,85){\vector(-966,-259){20}}
     \put(1,40){\vector(966,259){20}}
     \end{overpic}
     \caption{UNet$_{2D}$}
     \end{subfigure}
     \begin{subfigure}[b]{0.16\textwidth}
     \begin{overpic}[width=\textwidth, trim={7cm 8cm 7cm 7cm},clip]{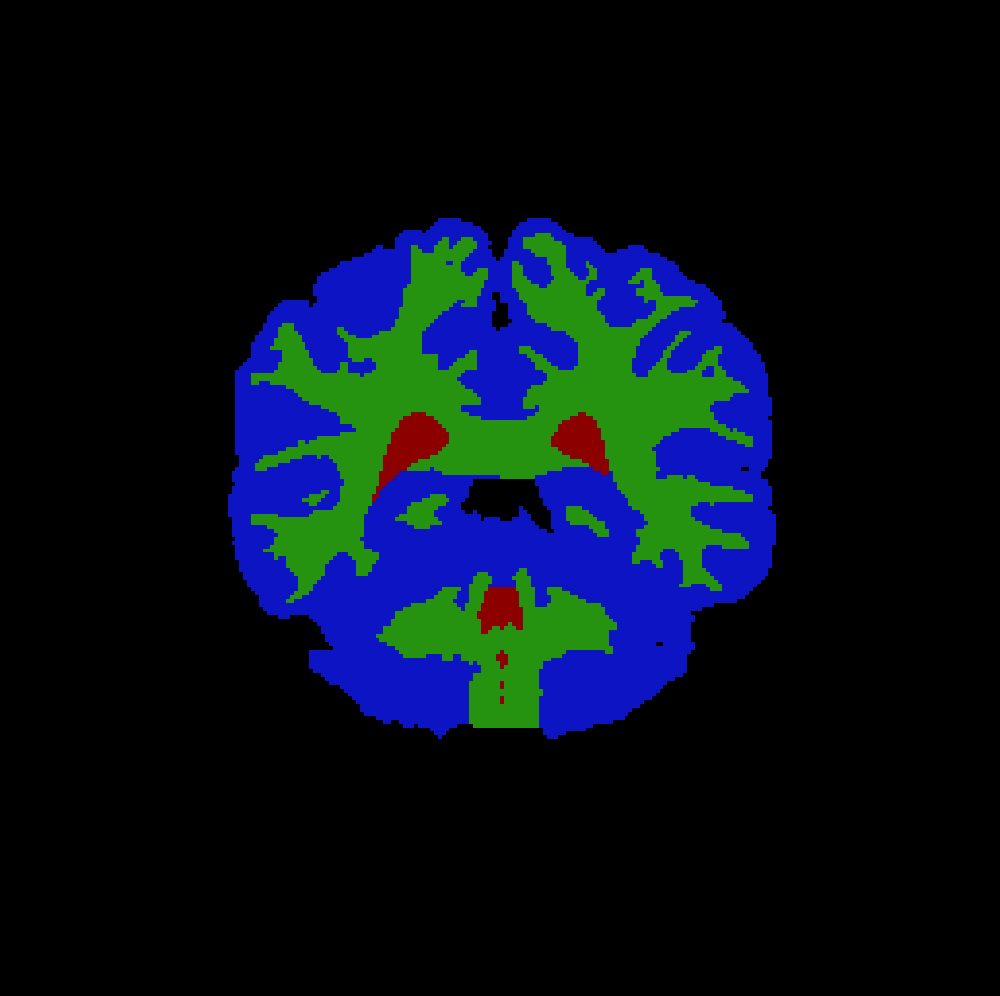}
     \linethickness{0.5mm} \color{white}%
      \put(90,85){\vector(-966,-259){20}}
     \put(1,40){\vector(966,259){20}}
     \end{overpic}
     \caption{UNet$_{3D}$}
     \end{subfigure}
    \caption{Segmentation output of the eight considered methods. The ground truth is displayed in (a) and the corresponding segmentation in (b-i). The colours for CSF, GM and WM, are red, blue and green, respectively. White arrows point out areas where differences w.r.t. the ground truth in (a) are more noticeable. Architectures using 2D information were trained with axial slices. \label{fig:qualitative-ibsr}}
\end{figure*}

Segmentation outputs obtained by the different methods on one of the volumes of the IBSR18 dataset are displayed in Fig.~\ref{fig:qualitative-ibsr}. Note that architectures using 2D information were trained with axial slices. As it can be seen in the illustration, since 2D architectures process each slice independently, the final segmentation is not necessarily accurate nor consistent: (i) subcortical structures exhibit unexpected shapes and holes, and (ii) sulci and gyri are not segmented finely. Thus, even if segmentation was carried out slice-by-slice, 3D approaches exhibit a smoother segmentation presumably since these methods exploit the 3D volumes directly. 

Another thing to note in Fig.~\ref{fig:qualitative-ibsr}f is that segmentation provided by KK$_{3D}$ seems worse than the rest -- even than its 2D analogue. The problem does not appear to be related to the number of parameters since KK$_{3D}$ requires the least amount of parameters in the 3D group, according to Table~\ref{tab:parameters}. This issue may be a consequence of the architectural design itself. Anisotropic voxels and heterogeneous spacing may be affecting the low-resolution path of the network considerably. Hence, the overall performance is degraded. 

In comparison with state of the art, our methods showed similar or enhanced performance. First, the best DSC scores for IBSR18 were collected by \cite{valverde2015comparison}. The highest values for CSF, GM and WM were $0.83\pm 0.08$, $0.88\pm0.04$ and $0.81\pm0.07$; while our best approach scored $0.90\pm0.03$, $0.96\pm0.01$ and $0.92\pm0.02$, for the same classes. Second, the best-known values for tissue segmentation using the MICCAI 2012 dataset, were reported by \cite{moeskops2016automatic}. Their strategy -- a multi-path CNN -- obtained $0.85\pm 0.04$ and $0.94\pm0.01$ for CSF and WM, respectively; while our best approach yielded $0.92\pm 0.03$ and $0.95\pm 0.02$. In this case, we cannot establish a direct comparison of GM scores since in Moeskops' case this class was subdivided into (a) cortical GM and (b) basal ganglia and thalami. Third, based on the results displayed in Table~\ref{tab:comparison}, our pipeline using DM$_{3D}$ led to the best segmentation results on the iSeg2017 leave-one-out cross-validation. Hence, we submitted our approach to the online challenge under the team name ``nic\_vicorob''\footnote{Results can be viewed at http://iseg2017.web.unc.edu/rules/results/}. The mean DSC values were $0.951$, $0.910$ and $0.885$ for CSF, GM and WM, correspondingly; and we also ranked top-5 in six of the nine evaluation scenarios (three classes, three measures).

\section{Discussion\label{sec:discussion}}
In this paper, we analysed quantitatively eight FCNN architectures inspired by the literature of brain segmentation related tasks. The networks were assessed through three experiments studying the importance of (i) overlapping patch extraction, (ii) multiple modalities, and (iii) network's dimensionality. 

Our first experiment evaluated the impact of overlapping as sampling strategy at training and testing stages. This overlapping sampling is explored as a workaround to the commonly used data augmentation techniques. This procedure can be used in this case as none of these networks processes a whole volume at a time, but patches of it. Based on our results, the technique proved beneficial since most of the networks obtained significantly higher values than when not considered. In particular, the four u-shaped architectures exhibited remarkable influence of this approach, presumably since more samples are used during training and since the same area is seen with different neighbouring regions, enforcing spatial consistency. Overlapping sampling in testing acts as a de-noising technique. We observed that this already-incorporated tool led to better performance than when absent as it helps filling small holes in areas expected homogeneous. The improvement was found to be at least $1$\%. Naturally, the main drawback of this technique is the expertise of the classifier itself, since it may produce undesired outputs when poorly trained.

Our second experiment assessed the effect of single and multiple imaging sequences on the final segmentation. We observed that regardless of the segmentation network, the inclusion of various modalities led to significantly better segmentations that in the single modality approach. This situation may be a consequence of networks being able to extract valuable contrast information. Improvements were noted concerning the mean as well as the dispersion of the values yielded by the methods. Although this outcome is aligned with the literature~\citep{Zhang2015}, additional trials on more datasets should be carried out to drawn stronger conclusions. Additionally, 
future work should consider evaluating tissue segmentation in the presence of pathologies and using more imaging sequences such as FLAIR and PD.

Our third experiment evaluated significant differences between 2D and 3D methods on the three considered datasets. In general, 3D architectures produced significantly higher performance than their 2D analogues. However, in one of our datasets, IBSR18, the results were quite similar between the two groups. Since the other two sets were re-sampled to obtain isotropic voxels, this outcome seems to be a consequence of the heterogeneity of the data in IBSR18, i.e. 2D methods seem to be more resilient to this kind of issues than 3D ones. 

Regarding network design, we observed that networks using information from shallower layers in deeper ones achieved higher performance than those using features directly from the input volume. Note the difference is intensified in heterogeneous datasets, IBSR18, where the latter strategy performs worse on average. Although it is only two networks, KK$_{2D}$ and KK$_{3D}$, this situation may underline the importance of internal connections (e.g. skip connections, residual connections) and fusion of multi-resolution information to segment more accurately. No remarkable difference was seen between convolutional-only and u-shaped architectures, except for processing times. In both training and testing, u-shaped networks produce segmentation faster than convolutional-only networks: u-shaped models require extracting less number of patches and provide a more prominent output at a time. For instance, in testing time and using a high degree of overlap, UNet$_{3D}$ can process a volume of size $256\times 256\times 256$ in around $130$ seconds while DM$_{3D}$ can take up to $360$ seconds.

Regarding general performance, two methods, DM$_{3D}$ and UNet$_{3D}$, displayed the best results. It is important to remark that our specific implementation of the latter architecture (i) required $30$\% fewer parameters to be set than the former, and (ii) classifies $\approx 32$K voxels more at a time which makes it appealing regarding computational speed. If the priority is time (training and testing), UResNet is a suitable option since it produces the same output size as the UNet, but the number of parameters is approximately half. Therefore, patch-based u-shaped networks are recommended to address tissue segmentation compared to convolutional-only approaches.

Taking into account results reported in the literature, we achieved top performance for IBSR18, MICCAI2012 and iSeg2017 with our implemented architectures. Three relevant things to note in this work. First, none of these networks has explicitly been tweaked to the scenarios, a typical pipeline has been used. Hence, it is possible to compare them under similar conditions. Approaches expressly tuned for challenges may win, but it does not imply they will work identically -- using the same set-up -- on real-life scenarios. Second, although these strategies have shown acceptable results, more development on domain adaptation and transfer learning (zero-shot or one-shot training) should be carried out to implement them in medical centres. Third, we do not intend to compare original works. Our implementations are inspired by the original works, but general pipelines are not taken into account in here. In short, our study focus on understanding architectural strengths and weaknesses of literature-like approaches.

\section{Conclusions\label{sec:conclusions}}
In this paper, we have analysed quantitatively $4\times 2$ FCNN architectures, 2D and 3D, for tissue segmentation on brain MRI. These networks were implemented inspired by four recently proposed networks~\citep{cciccek20163d, Dolz2017, guerrero2017white, kamnitsas2017efficient}. Among other characteristics, these methods comprised (i) convolutional-only and u-shaped architectures, (ii) single- and multi-modality inputs, (iii) 2D and 3D network dimensionality, (iv) varied implementation of multi-path schemes, and (v) different number of parameters.

The eight networks were tested using three different well-known datasets: IBSR18, MICCAI2012 and iSeg2017. These datasets were considered since they were acquired with different configuration parameters. Testing scenarios evaluated the impact of (i) overlapping sampling on both training and testing, (ii) multiple modalities, and (iii) 2D and 3D inputs on tissue segmentation. First, we observed that extracting patches with a certain degree of overlap among themselves led consistently to higher performance. The same approach on testing did not show a relevant improvement (around $1$\% in DSC). It is a de-noising tool that comes along with the trained network. Second, we noted that using multiple modalities -- when available -- can help the method to achieve significantly higher accuracy values. Third, based on our evaluation for tissue segmentation, 3D methods tend to outperform their 2D counterpart. However, it is relevant to recognise that 3D methods appear slightly more affected to variations in voxel spacing. Additionally, networks implemented in this paper were able to deliver state-of-the-art results on IBSR18 and MICCAI2012. Our best approach on the iSeg2017 ranked top-5 in most of the on-line testing scenarios.

To encourage other researchers to use the implemented evaluation framework and FCNN architectures, we have released a public version of it at our research website.

\section*{Acknowledgments}
Jose Bernal and Kaisar Kushibar hold FI-DGR2017 grants from the Catalan Government with reference numbers 2017FI B00476 and 2017FI B00372 respectively. Mariano Cabezas holds a Juan de la Cierva - Incorporación grant from the Spanish Government with reference number IJCI-2016-29240. This work has been partially supported by La Fundació la Marató de TV3, by Retos de Investigació TIN2014-55710-R, TIN2015-73563-JIN and DPI2017-86696-R from the Ministerio de Ciencia y Tecnolog\'ia. The authors gratefully acknowledge the support of the NVIDIA Corporation with their donation of the TITAN-X PASCAL GPU used in this research. The authors would like to thank the organisers of the iSeg2017 challenge for providing the data.

\appendix

\bibliography{mybibfile}

\end{document}